\documentclass[conference]{IEEEtran}
\IEEEoverridecommandlockouts
\usepackage{algorithm}
\usepackage{algpseudocode}
\usepackage{graphicx}
\usepackage{textcomp}
\usepackage{enumitem}
\usepackage{multirow}
\usepackage{subfigure}
\usepackage{amsmath,bm}
\usepackage{xcolor}
\usepackage{url}
\usepackage{hyperref}
\hypersetup{hidelinks}
\usepackage{gensymb}

\newcommand{\CommentZheng}{}
\newcommand{\newCommentZheng}{}
\newcommand{\nnCommentZheng}{}
\newcommand{\newadd}{}

\def\BibTeX{{\rm B\kern-.05em{\sc i\kern-.025em b}\kern-.08em
    T\kern-.1667em\lower.7ex\hbox{E}\kern-.125emX}}
\begin{document}

\title{On Inferring User Socioeconomic Status with Mobility Records}

\author{\IEEEauthorblockN{Zheng Wang\IEEEauthorrefmark{1}\IEEEauthorrefmark{2},
Mingrui Liu\IEEEauthorrefmark{1},
Cheng Long\IEEEauthorrefmark{1}\IEEEauthorrefmark{5}\IEEEcompsocitemizethanks{\IEEEcompsocthanksitem\IEEEauthorrefmark{5}Corresponding author.},
Qianru Zhang\IEEEauthorrefmark{3},
Jiangneng Li\IEEEauthorrefmark{1},
Chunyan	Miao\IEEEauthorrefmark{1}\IEEEauthorrefmark{4}
}
\IEEEauthorblockA{\IEEEauthorrefmark{1}School of Computer Science and Engineering, Nanyang Technological University, Singapore,\\
\IEEEauthorrefmark{2}Huawei Singapore Research Center, Singapore,\\
\IEEEauthorrefmark{3}Department of Computer Science, The University of Hong Kong, Hong Kong SAR\\
\IEEEauthorrefmark{4}China-Singapore International Joint Research Institute (CSIJRI), China\\
\{zheng011,mingrui001,jiangnen002\}@e.ntu.edu.sg, \{c.long,ascymiao\}@ntu.edu.sg, qrzhang@cs.hku.hk}}

\maketitle

\begin{abstract}
When users move in a physical space (e.g., an urban space), they would have some records called mobility records (e.g., trajectories) generated by devices such as mobile phones and GPS devices. Naturally, mobility records capture essential information of how users work, live and entertain in their daily lives, and therefore, they have been used in a wide range of tasks such as user profile inference, mobility prediction and traffic management. In this paper, we expand this line of research by investigating the problem of inferring user socioeconomic statuses (such as prices of users' living houses as a proxy of users' socioeconomic statuses) based on their mobility records, which can potentially be used in real-life applications such as the car loan business. 
For this task, we propose a socioeconomic-aware deep model called \texttt{DeepSEI}. The \texttt{DeepSEI} model incorporates two networks called deep network and recurrent network, which extract the features of the mobility records from three aspects, namely spatiality, temporality and activity, one at a coarse level and the other at a detailed level. We conduct extensive experiments on real mobility records data, POI data and house prices data.
The results verify that the \texttt{DeepSEI} model achieves superior performance than existing studies.
All datasets used in this paper will be made publicly available. 
\end{abstract}


\section{INTRODUCTION}
\label{sec:introduction}

With the rapid development of GPS devices and mobile technologies, recent years have witnessed an unprecedented growth in mobility data. This big amount of data has attracted many research efforts to acquire knowledge of human mobility behaviors. More specifically, extensive studies have been conducted on profiling users from mobility records. For example, it has been explored to infer users' demographic attributes from their check-ins~\cite{zhong2015you}, users' ethics and gender from their photo sharing data with geo tags~\cite{riederer2020location}, passengers' employment statuses from their smart card data~\cite{zhang2019deep, ding2019estimating}, and users' demographics from their trajectories~\cite{wu2019inferring}, etc.
%
While these techniques are extensive and have some merits, there still exist some scenarios that have been overlooked and/or cannot be adequately solved by them. 
For example, in
some real-life applications such as car loans, 
quite many demographic attributes such as the age and gender of the users are already provided by users.
What is demanded for these applications is to infer the socioeconomic statuses of users, e.g., the prices of their living houses and whether they will pay their monthly loans on time, etc.
Yet these have been mostly overlooked by existing studies~\cite{zhong2015you,riederer2020location,zhang2019deep,ding2019estimating,wu2019inferring}.

In this paper, we aim to infer users' socioeconomic statuses from their mobility records.
This is motivated by two considerations. 
First, users' socioeconomic statuses are closely linked to where they live or work, both of which could be potentially reflected by their mobility records. Second, users' socioeconomic statuses can sometimes be disclosed by the places they visit, especially those they visit during weekends, and the patterns of their visits at these places, which again could be revealed by their mobility records.
Here, a user's socioeconomic status can refer to many different indicators, such as the price range of the user's living house~\cite{xu2018human,ding2019estimating}, the likelihood that the user will pay a car loan installment on time, or the user's income, etc. Constrained by the availability of datasets and privacy concerns, in this paper, we infer the home location of a user based on his/her mobility records (i.e., Geolife) and then crawl the house price data from the Web based on the home location as the proxy of the user's socioeconomic status. 
Since both the mobility records data and the house price data are publicly available, no privacy will be broken in this study.

%
Specifically, we propose a socioeconomic-aware deep model called \texttt{DeepSEI} for user socioeconomic status inference. 
In \texttt{DeepSEI}, it first preprocesses the users' mobility records data by filtering the noises, extracting the stay points, and inferring the activities behind the extracted stay points.
Then, it incorporates two networks, namely \emph{deep network} and \emph{recurrent network}, to capture users' activities data at a coarse level and at a detailed level, respectively, for this task. 
The deep network aims to capture some statistics based on users' mobility records (i.e., at a coarse level) and the recurrent network aims to capture the sequential patterns behind users' mobility records (i.e., at a detailed level).

The deep network takes as inputs three features of users' mobility records data, including spatiality diversity, temporality diversity and activity diversity. 
Spatiality diversity captures the spatial information in the territory where users' daily activities are conducted. 
Temporality diversity captures the temporal regularity of users, which can potentially help to indicate their professions, e.g., self-employers tend to stay at home and only go out occasionally, while some users working at a government department would commute more regularly. 
Activity diversity reveals the diversity of movements among users' activity locations, which can reflect users' socioeconomic statuses as shown in~\cite{xu2018human,wu2019inferring}.


The recurrent network takes as inputs the sequences of activities of users, where each activity has spatial, temporal and semantic features. The spatial and temporal features indicate where and when the activities are conducted and the semantic features, e.g., working or shopping, indicate the activity types and provide the context for understanding users' daily routines.
%
We adopt a hierarchical LSTM with two levels for the recurrent network.
The activities within a day are modeled in the low-level LSTM and the activities within days are modeled in the high-level LSTM. The hierarchical LSTM brings two advantages. First, users' sequences of mobility records are generally long, and the hierarchical structure can reduce the length and alleviate the issue of degraded performance for handling long sequences. Second, users' mobility records are organized on a daily basis, and the two-level LSTMs preserve the users' periodic information. 
{\newadd{We note that other sequence encoder models, e.g., Transformer, are also applicable for the task, and we leave it a future work to explore these models.}}

The novelty of the paper is two-fold. First, the problem setting is new. To our best knowledge, there are no existing studies that take GPS mobility records as inputs and infer users’ socioeconomic statuses. 
Second, our method is distinctive from existing ones in that it explores a data-driven solution with two well-designed neural networks for the inference task.
%
In summary, we make the following contributions: 
\begin{itemize}[leftmargin=9pt]
    \item We study a novel problem of 
    inferring users' socioeconomic statuses based on their GPS mobility records. This problem is new and has practical applications in real life (e.g., risk assessment for car loan applications/managements).
    \item We propose a novel learning framework called \texttt{DeepSEI} for the problem, which is a supervised deep learning model and incorporates two neural networks (i.e., deep network and recurrent network) to capture the features from three aspects of users' mobility records, i.e., spatiality, temporality and activity, at both coarse and detailed levels.
    
    \item We conduct the experiments on real-world GPS trajectory, POI and house price datasets, which are publicly available, and the results demonstrate our method's superior performance for the task, e.g., our method outperforms the best baseline by at least 15\% in terms of prediction accuracy. 
\end{itemize}


\section{RELATED WORK}
\label{sec:related}


\subsection{Human Mobility Analytics}
Earlier studies~\cite{hanson1981travel,kwan1999gender,limtanakool2006influence} examine the relationships between human travel behaviors with their profiles, including gender~\cite{kwan1999gender,limtanakool2006influence}, age~\cite{limtanakool2006influence,kwan1999gender}, race~\cite{kwan1999gender}, employment status and income~\cite{hanson1981travel}. For example, Hanson et al.~\cite{hanson1981travel} suggest that an individual's employment status and income have a positive impact on his/her travel frequency. Kwan et al.~\cite{kwan1999gender} reveal that men are more inclined to visit recreation places than women. The earlier studies make a great impact on the subsequent research; however, the works are conducted on travel survey data, which needs to be collected manually, and consequently the findings are mainly based on a small number of volunteers for a short period of time.

With the proliferation of communication techniques, mobile phone data becomes a new data source for conducting this type of research. Mobile phone data is collected from cellphone users, where the mobile phone records track a user's id, the user's location when he/she makes a phone call (the location is reported as the longitude and latitude), and the timestamp at which the phone call starts. With the mobile phone data, Xu et al.~\cite{xu2015understanding} present a home-based approach to analyze
human activities in Shenzhen. They find that people who live in the northern part of Shenzhen are generally with a small activity space around their homes, and people with a larger activity space mainly live in the southern part, where the economy is highly developed. Further, they propose an analytical framework for understanding the relationships between human mobility and socioeconomic status~\cite{xu2018human}. Specifically, they take two cities, Singapore and Boston, for case studies, and reveal an interesting finding that the richer tend to travel shorter in Singapore but longer in Boston. 
Blumenstock et al.~\cite{blumenstock2015predicting} predict individual socioeconomic status (e.g., poverty and wealth levels) with users' survey data collected from their historical mobile calls. 
In addition, Huang et al.~\cite{huang2016activity} explore possible factors that may influence individual daily activities, and the results demonstrate that socioeconomic status, urban spatial structure, work place and region geographical layout all play a critical role. Kelly et al.~\cite{kelly2013uncovering} use the location data collected from mobile sensors to identify some predictability patterns that can be linked to users' demographics such as age, gender and social meeting contacts, etc.
{\newadd{Ding et al.~\cite{ding2019estimating} estimate users’ socioeconomic statuses from their subway smart card data, which records users' pick-up and drop-off locations at subway stations for each trip.}}
Different from these studies, we propose to infer users' socioeconomic statuses with their GPS trajectories and develop a deep learning based model called \texttt{DeepSEI}.

\subsection{Mobility and/or Temporality Prediction}
We review the existing studies regarding the prediction task, where mobility and temporality information is involved. For mobility prediction, with the proliferation of location-based services, it has been a hot research topic in recent years. Mobility prediction aims at predicting the next location for the user, while POI recommendation aims to predict the following several locations that the user will visit. 
Earlier studies~\cite{monreale2009wherenext, zhang2014splitter} for the next location prediction task are based on exacting the historical user mobility patterns. In recent years, many learning-based methods~\cite{chen2020context,ju2020interaction,feng2018deepmove} are proposed to model the users' mobility patterns in a data-driven manner. For example, DeepMove~\cite{feng2018deepmove} is an attentional recurrent neural network based model to capture the user's periodical patterns for his/her mobility prediction.
%
%
Chen et al.~\cite{chen2020context} propose a context-aware deep model called DeepJMT to jointly predict where and when a user will visit next, which considers both users' visit histories and the spatial and user contexts of the visits.
For temporality prediction, many existing studies~\cite{xiao2017modeling,du2016recurrent} adopt temporal point process to model the time as a sequence of discrete random events, and then jointly predict the next event type and timestamp. 
Our problem differs from these studies mainly in that we aim to predict users' socioeconomic statuses but not their mobility and/or temporality. 

\section{Problem Description}
\label{sec:preliminary}

In this paper, we aim to infer users' socioeconomic statuses from their mobility records.
%
%
Here, a user's socioeconomic status can refer to many different indicators, such as the price range of the user's living house~\cite{xu2018human,ding2019estimating}, the likelihood that the user will pay a car loan installment on time, or the user's income, etc. Constrained by the availability of datasets and privacy concerns, in this paper, we infer the home location of a user based on his/her mobility records (i.e., Geolife) and then crawl the house price data from the Web based on the home location as the indicator of the user's socioeconomic status. 
{\newadd{Many studies~\cite{xu2018human,ding2019estimating,mohamed2016clustering} show that the house price data reflects users' socioeconomic statuses well. For example, in~\cite{xu2018human}, it reveals that a user's house price is strongly correlated with a user's monthly income.}}
More specifically, we model the socioeconomic status inference task as a multi-class classification task, where each class corresponds to a range of house prices. In experiments, we vary the number of classes from 2 to 5.
Since both the mobility records data and the house price data are publicly available, no privacy will be broken.

\section{METHODOLOGY}
\label{sec:method}

\subsection{Overview}
To predict users’ socioeconomic statues with their mobility records, we propose a novel socioeconomic-aware deep model 
called \texttt{DeepSEI}. \texttt{DeepSEI} model consists of three components, including data preprocessing (Section~\ref{sec:preprocess}), deep network (Section~\ref{sec:static}) and recurrent network (Section~\ref{sec:dynamic}).
The data preprocessing component is to filter the noises in the mobility records data, extract the stay points from the mobility records, and further infer the activities behind the extracted stay points.
The deep network aims to capture some statistics based on users' mobility records (i.e., at a coarse level) and the recurrent network aims to capture the sequential patterns behind users' mobility records (i.e., at a detailed level). Therefore, the two networks collaboratively capture rich information from users' mobility records data.
\begin{figure}
  \centering
  \includegraphics[width=\linewidth]{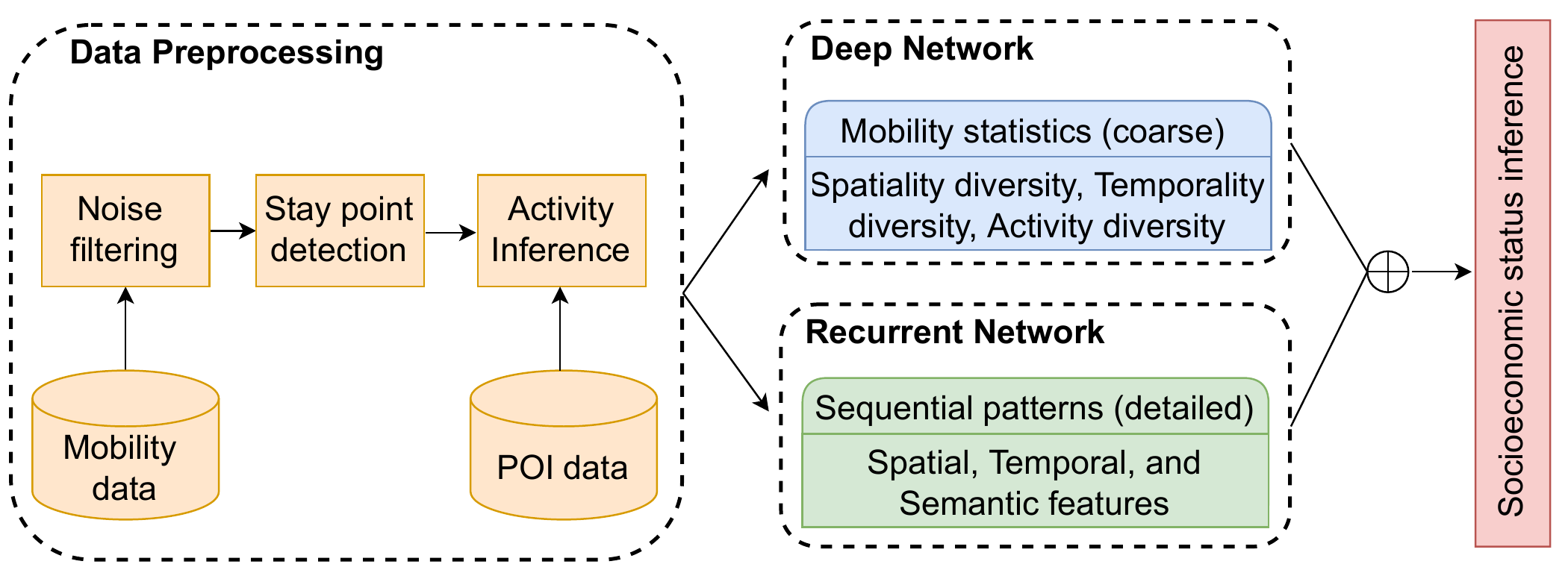}
  \caption{The overall framework of \texttt{DeepSEI}, where $\bigoplus$ denotes the concatenation operation.}
  \label{fig:model}
  \vspace{-6mm}
\end{figure}

{\CommentZheng{
For the deep network, it is designed to quantify three important aspects of users' mobility characteristics, namely spatiality, temporality and activity. 
For spatiality, we consider users' radii of gyration extracted from the their trajectories, which describe the typical spatial range of users' activities.
For temporality, it is intuitive that people with different professions (e.g., government officials or IT engineers) would travel with different temporal patterns/regularities of activities. 
We explore an entropy-based temporality indicator for capturing the regularity of users' activities. 
For activity, it is intuitive that an individual's daily activities could be used for inferring the user's socioeconomic status, e.g., rich people have diverse daily activities and travel more places in general, and thus we use an entropy-based activity indicator to capture that aspect. 

For the recurrent network, we focus on the sequential activity data extracted from users' mobility records, where 
each activity involves spatial, temporal, and semantic features. 
In particular, spatial and temporal features record where and when the user's activities are conducted, respectively. Semantic feature provides the context for understanding the intended activities that a user conducts, e.g., working or dining. Intuitively, the contextual semantics would be useful to profile the user's socioeconomic status, e.g., if a user visits the fast food restaurants frequently and rarely visits recreational venues such as fitness centers, he may have poor socioeconomic conditions. }}
We adopt a hierarchical LSTM for the recurrent network since the hierarchical structure can reduce the mobility sequence length, and extract periodic mobility behaviours of human.

We concatenate the outputs of the two networks into a long vector, which is further fed into a fully connected layer to produce the predicted users' statuses. 
We train the \texttt{DeepSEI} model in a supervised manner (Section~\ref{sec:training}). Figure~\ref{fig:model} illustrates the overall framework of \texttt{DeepSEI}. Next, we discuss the detailed designs of each component.
 
\subsection{Data Preprocessing}
\label{sec:preprocess}
The raw mobility records data corresponds to a sequence of sampled locations with time stamps. The data may involve noises (e.g., the GPS noises) and lack of semantics (e.g., activities). 
The data preprocessing component is to preprocess the data by (1) filtering the noises, (2) extracting the stay points (which indicate activity behaviors), and (3) inferring the semantics of the stay points (e.g., the categories of the POIs that are probably visited at the extracted stay points).
%
Specifically, data preprocssing involves three steps, namely Noise Filtering, Stay Point Detection, and Activity Inference.

\smallskip
\noindent \textbf{Noise Filtering.} For mobility records data (or trajectory data), we first perform trajectory segmentation by dividing the data into one-week trajectory instances, which are used to train and test the model.
Within each segmented trajectory, we filter those noisy data points and then detect those stay points, where a stay point corresponds to one activity such as working, shopping, and staying at home. In particular, trajectory data usually contains noises due to the way how it is collected, e.g., a sampled location might be several hundred meters away from a true location. Such noisy points will affect the quality of stay point detection. We adopt a heuristic-based approach proposed in~\cite{zheng2015trajectory} for filtering noisy points in trajectories. It sequentially calculates the traveling speed for each point in a trajectory based on its precursor and itself. If the speed is larger than a threshold, the current examined point is removed from the point sequence.

\smallskip
\noindent \textbf{Stay Point Detection.} Based on the cleaned trajectories, we extract all stay points from them. Specifically, we adopt the stay point detection algorithm proposed in~\cite{li2008mining} for detecting stay points from trajectories. The algorithm first checks if the distance between an anchor point and its successors in a trajectory is larger than a given threshold $S_d$. It then calculates the duration between the anchor point and the last successor within $S_d$. If the duration is larger than a temporal threshold $S_t$, a stay point is detected, and the anchor point moves to the next point after the current stay point. Otherwise, the anchor point moves forward by one. This process is repeated until the anchor point moves to the end of the sequence so that all stay points in a trajectory are extracted.

\begin{figure}
\centering
\begin{tabular}{c}
   \begin{minipage}{0.9\linewidth}
    \includegraphics[width=\linewidth]{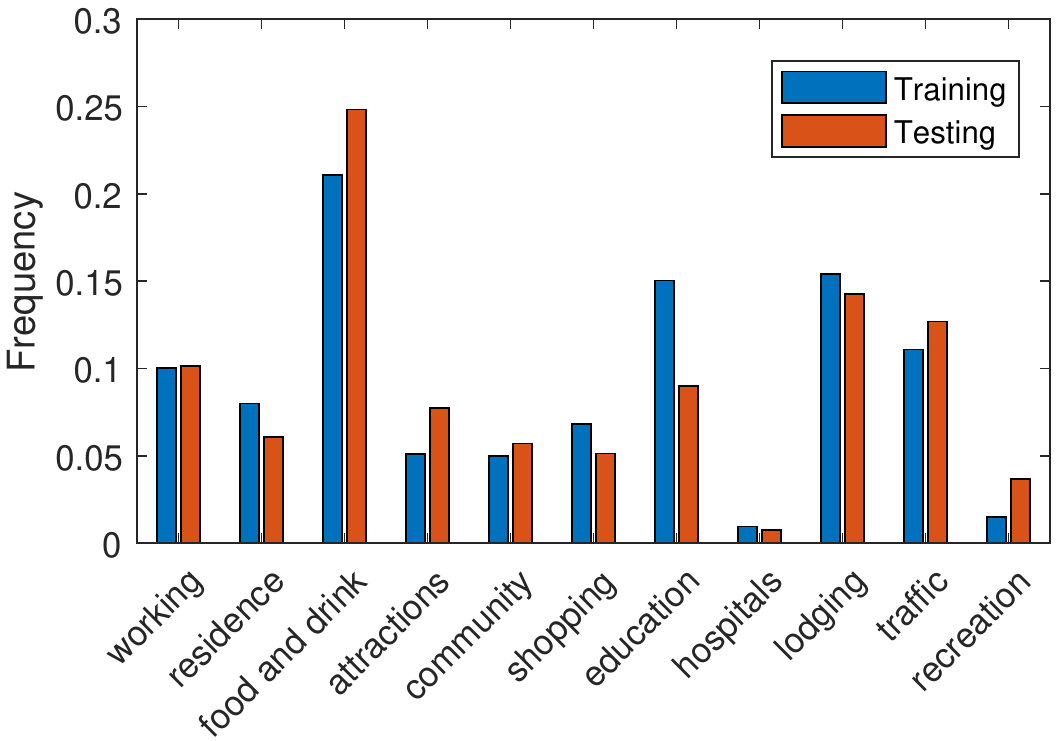}
    \end{minipage}
\end{tabular}
\vspace*{-2mm}
\caption{Activity distribution of Beijing.}
\label{fig:poi}
\vspace*{-4mm}
\end{figure}

\smallskip
\noindent \textbf{Activity Inference.} Based on the stay points extracted from users' mobility records, we infer the most relevant activity for each stay point. 
For each stay point, we check 8 neighbouring grid cells of the grid cell, in which the stay point is located. Then, we infer the activity associated with the stay point to be the most frequent POI category among those POIs in the 8 neighbouring grid cells.  In addition, if no POI is found within the grids, we infer the activity to be a special tag called ``other''. 
In Figure~\ref{fig:poi}, we illustrate the inferred activity distribution for Beijing in terms of training and testing (details will be presented in the sequel). We notice some other methods such as Markov based inference models~\cite{wu2016did,yan2013semantic,yan2011semitri} are also applicable for the task of activity inference. Since the POI categories are in the form of discrete tokens, we obtain the activity vectors {\CommentZheng{by embedding their tokens as one-hot vectors.}}

\subsection{Deep Network}
\label{sec:static}
One previous study~\cite{xu2018human} reveals that users' socioeconomic statuses can be reflected by their mobility patterns/statistics. 
Inspired by this, we adopt some indicators that are computed from users' mobility records for our task. These indicators capture the spatiality, temporality and activity aspects of users' mobility patterns. 
We embed indicators via a deep network and concatenate the embeddings for our task. 

\smallskip
\noindent \textbf{Spatiality Diversity.}
To capture the mobility features from users' trajectory data, we consider radius of gyration, which is widely used as a spatial indicator to capture users' mobility characteristics~\cite{xu2018human,wu2019inferring}. Given a trajectory data $T = <(p_1,t_1),(p_2,t_2),...,(p_n,t_n)>$, where $p_i$ and $t_i$ ($1 \le i \le n$) denote the location $p_i$ of a moving object at time $t_i$. The radius of gyration $R_g$ is defined as follows.
\begin{equation}
    R_g = \sqrt{\frac{\sum_{i=1}^{n}(p_i-p_c)^2}{n}}, \quad p_c=\frac{\sum_{i=1}^{n}p_i}{n},
\end{equation}
where $p_c$ denotes the center of the sampled locations. The rationale of radius of gyration is to capture the spatial dispersion of a user' movement. Intuitively, a small radius indicates that the user's activities are mainly in a small area.

\smallskip
\noindent \textbf{Temporality Diversity.} We consider the feature in temporal aspect. It is intuitive that people with similar professions would travel similarly in terms of temporality. For example, office staff would commute between home and office regularly on weekdays while self-employers would mainly stay at home and only go out occasionally. In addition, some users (e.g., those work at a government department) would commute more regularly than others (those work at an IT company). These temporality patterns embed rich information that could be used for inferring socioeconomic statuses of users. Therefore, we explore a temporality indicator called temporality diversity. 

Given a user's stay points extracted from his/her mobility records ${(s_1, \Delta t_1), (s_2, \Delta t_2),...,(s_n,\Delta t_n)}$, where $s$ and $\Delta t$ denote the stay location and the duration of staying at that location, respectively. By following~\cite{scheiner2014gendered,xu2018human}, we calculate the temporality diversity via cross-entropy, which captures the duration distribution among those stay locations. Let $pt_i=\frac{\Delta t_i}{\sum_{j=1}^{n}\Delta t_j}$ denote the proportion of stay duration at location $s_i$, the temporality diversity (TD) is defined as follows:
\begin{equation}
    TD = -\sum_{i=1}^{n}pt_i\log(pt_i),
\end{equation}
where $\sum_{i=1}^{n}pt_i = 1$.
The rationale of the indicator is to consider the human daily regularity. For example, for some people (e.g., IT programmers), whose daily lives are mainly concentrated at home and work places, they would have a high regularity reflected as a low cross-entropy value.

\smallskip
\noindent \textbf{Activity Diversity.}
Previous studies~\cite{pappalardo2015using, xu2018human} exhibit that the diversity of individual daily activities has a strong correlation with their socioeconomic statuses. Intuitively, a well-developed city (e.g., Beijing in our study) provides many facilities for residents to conducct various activities, and richer people tend to have a higher activity diversity in daily lives, e.g., their jobs are generally with higher diversification.

Inspired by this, we explore the features for describing activity diversity in the model. 
Given a sequence of user's stay points $(s_1, \Delta t_1), (s_2, \Delta t_2),...,(s_n,\Delta t_n)$, where each stay location $s$ is potentially 
associated with an activity such as working. 
We define two consecutive stay locations in a trip $e = (s_{i-1}, s_i) (0<i \leq n)$ as a source and destination pair. 
We let $E$ denote the set of all possible source and destination pairs extracted from the whole stay points, where the direction of movement could be ignored. For each pair $e \in E$, $p(e)$ denotes the proportion of observing the movement corresponding to the pair $e$ wrt the total number of movements (i.e., $n-1$) in the records. For example, given a user's stay points ${(a, \Delta t_1), (b, \Delta t_2), (c, \Delta t_3), (d, \Delta t_4), (c, \Delta t_5), (b, \Delta t_6), (a, \Delta t_7)}$, there are 6 movements in the records and 3 source and destination pairs in $E$ without considering direction, i.e., $(a,b)$, $(b,c)$ and $(c,d)$. For the pair $e=(a,b)$, $p(e)$ is calculated as $p(e)=\frac{2}{6}=0.33$. Note that $\sum_{e \in E}p(e)=1$, and we define the Activity Diversity (AD) via cross-entropy as follows:
\begin{equation}
    AD = -\sum_{e \in E}p(e)\log(p(e)).
\end{equation}
The rationale is that for a user, whose trips are concentrated on a few locations, he/she would have a high cross-entropy value.

\begin{figure}
  \centering
  \includegraphics[width=0.9\linewidth]{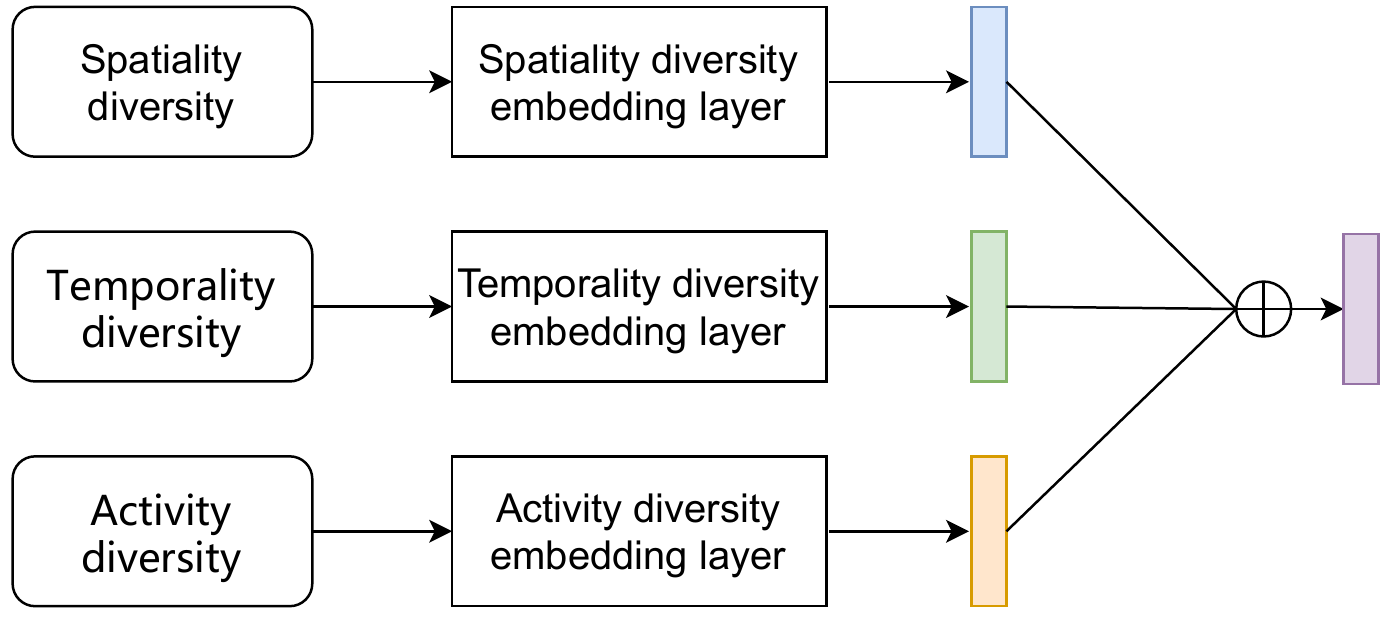}
  \vspace{-3mm}
  \caption{The architecture of deep network, where $\bigoplus$ denotes the concatenation.}
  \label{fig:model_static}
  \vspace{-5mm}
\end{figure}

\smallskip
\noindent \textbf{Network Architecture.}
Figure~\ref{fig:model_static} illustrates the architecture of the deep network. To feed the above features into the \texttt{DeepSEI} model, we first tokenize them. In particular, spatiality diversity, temporality diversity and activity diversity are naturally in the form of continuous float values, we tokenize them by partitioning the range with a predefined granularity, e.g., we partition the range 0.09-8,143.3
(resp. {\newCommentZheng{0-5.73 and 0.02-5.36}}) of spatiality diversity (resp. temporality diversity and activity diversity) with a 100 (resp. 0.5 and 0.5) granularity, and thus we obtain 82 (resp. 11 and 10) tokens.

After obtaining the tokens of above features, we then embed each feature via an embedding layer, and denote the embedded features (i.e., 32-dimensional vectors) to be fed to the \underline{d}eep network for \underline{s}patiality diversity, \underline{t}emporality diversity and \underline{a}ctivity diversity as $x^{ds}$,$x^{dt}$ and $x^{da}$, respectively. We concatenate the embedded feature vectors, and thus obtain a long vector with dimension $32*3$, which will be further used. To facilitate the training process, we pre-train the tokens of spatiality diversity, temporality diversity and activity diversity via Skip-Gram model~\cite{mikolov2013efficient}. After pre-training, the tokens with similar context (e.g., two tokens correspond to two similar values for a feature) will be embedded into similar vectors in a latent space. We then use the pre-trained vectors to initialize the embedding layer of the deep network, and those embeddings can further be optimized with the model training.

\subsection{Recurrent Network}
\label{sec:dynamic}
The recurrent network aims to capture sequential patterns of users' activities, each of which involves spatial, temporal and semantic features.
We first embed these features of each activity and then feed the embeddings to a hierarchical LSTM network.
Compared with the implementation of vanilla RNN based models, hierarchical LSTM is more capable of capturing the sequential and periodic information from users' trajectory data. 

\smallskip
\noindent \textbf{Spatial Embedding.} 
We partition the geographical space into grid cells with the grid size of 200m $\times$ 200m, and map the GPS coordinates of the stay point behind an activity to a grid cell. Naturally, the mapped grid cell, which encompasses the coordinates of the stay point, can represent its spatial information. We then obtain the spatial embedding of the activity as {\CommentZheng{a one-hot vector}}, which will be further fed into the hierarchical LSTM.

\smallskip
\noindent \textbf{Temporal Embedding.} For the extracted stay point behind each activity, it is associated with a timestamp to record the starting time of the stay. We take the timestamp as a temporal feature, and embed it into the model. 
By following the previous study~\cite{zhong2015you}, we split a one-week trajectory data into two parts, i.e., one for weekdays and one for weekends. For each part, we further split a day into 24 hourly time bins, and therefore we obtain $24*2=48$ time bins in total to tokenize the temporal feature, where we take the temporality for weekdays and weekends differently. This is because users generally have different lifestyles for these two parts. {\CommentZheng{Similarly, we obtain the temporality vectors by embedding their time bins as one-hot vectors}}, and fed them into the hierarchical LSTM.

\smallskip
\noindent \textbf{Semantic Embedding.}
The semantic feature of an activity corresponds to the POI we have inferred for a stay point (details in Section~\ref{sec:preprocess}). We obtain the embeddings of semantic features (i.e., POIs) as one-hot vectors.

\begin{figure}
\vspace{-3mm}
  \centering
  \includegraphics[width=\linewidth]{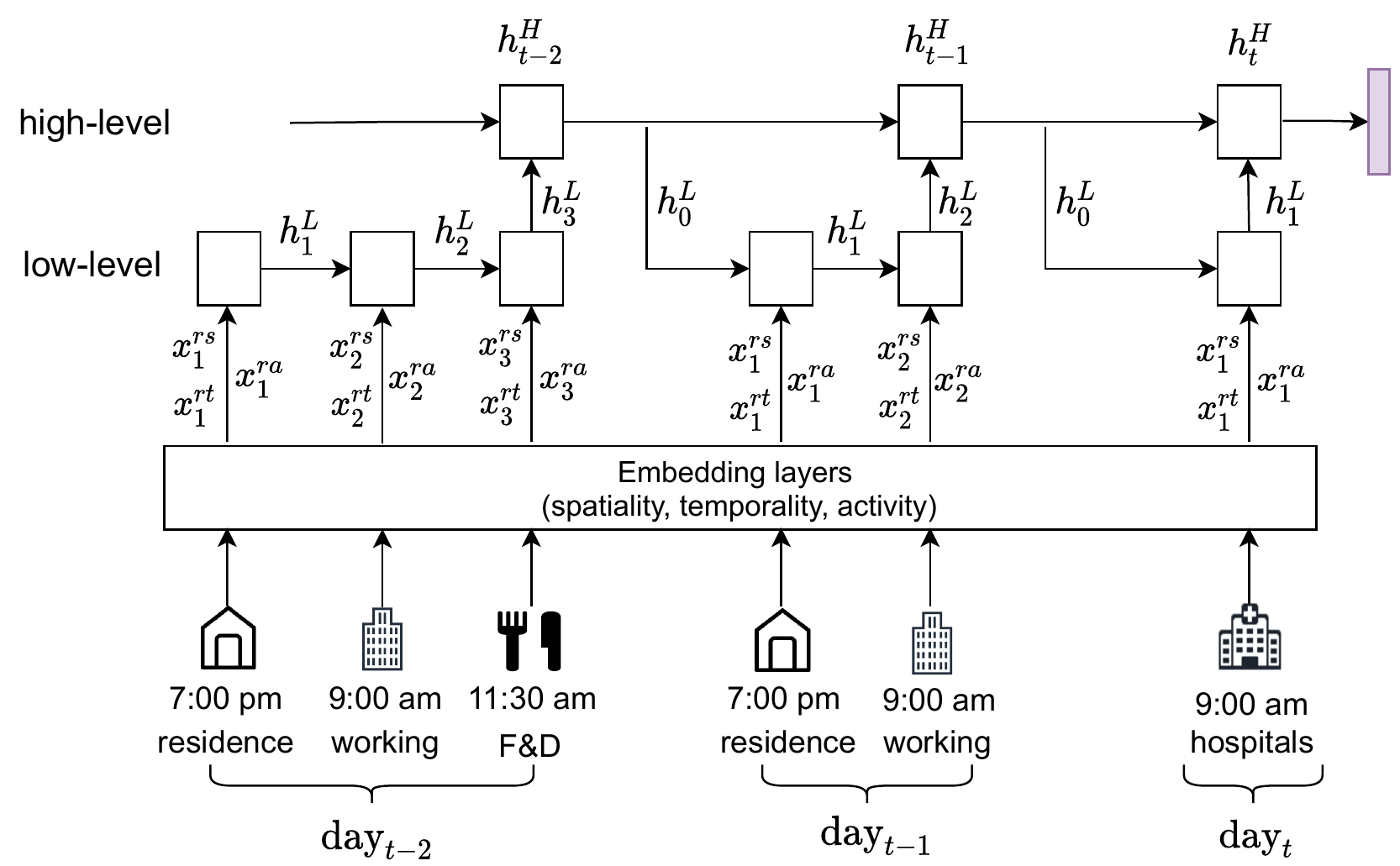}
  \vspace{-5mm}
  \caption{The architecture of recurrent network.}
  \label{fig:model_dynamic}
  \vspace{-5mm}
\end{figure}

\smallskip
\noindent \textbf{Network Architecture.}
Figure~\ref{fig:model_dynamic} illustrates the architecture of the recurrent network, which is implemented with a hierarchical LSTM, where the LSTM units~\cite{hochreiter1997long} are employed for both low-level and high-level structures. Take an one-week trajectory for example, the low-level LSTM is to capture the intra-transitions within a day and the high-level LSTM is to capture the inter-transitions across days.

For the low-level LSTM, we denote the embeddings used in the \underline{r}ecurrent network for \underline{s}patiality, \underline{t}emporality and \underline{a}ctivity as $x^{rs}$, $x^{rt}$ and $x^{ra}$. Then, we concatenate those embeddings at each time step $i$, denoted by $x^{rs}_i \oplus x^{rt}_i \oplus x^{ra}_i$, and feed the concatenation to low-level LSTM to obtain the hidden state $h^{L}_i$ at this time step.
\begin{equation}
h^{L}_i = \text{LSTM}^{L}(x^{rs}_i \oplus x^{rt}_i \oplus x^{ra}_i, h^{L}_{i-1}).
\end{equation}
We take the last hidden state within the day denoted by $h^{L}_n$, as a latent representation for capturing the intra-transitions within the day, and it is further fed into high-level LSTM. 
\begin{equation}
h^{H}_j = \text{LSTM}^{H}(h^{L}_n, h^{H}_{j-1}), 
\end{equation}
where $h^{H}_j$ denotes the high-level hidden state at the $j^{th}$ day, which is then fed to initialize the hidden state $h^{L}_0$ for the low-level LSTM. We feed the last hidden state at the high-level LSTM (i.e., suppose $h^{H}_7$ is the last day of a week) into a fully connected layer as the output of recurrent network.

With the hierarchical LSTM, it brings two advantages. First, compared with vanilla RNN, modeling the user’s mobility records in a hierarchical way is able to reduce the sequence length, where the low-level LSTM is for handling the records within a day and the high-level LSTM is for handling the records across days. The design helps alleviate the issue of degraded model performance for modeling long mobility sequences. Second, the hierarchical structure is able to capture both sequential and periodic information, where the low-level LSTM captures the sequential transitions from users’ mobility records, and high-level LSTM preserves the periodic information on a daily basis.

\smallskip
\subsection{Jointly Training deep and recurrent networks}
\label{sec:training}
We jointly train the two networks (i.e., deep and recurrent networks) for predicting users' socioeconomic statuses. 
Following the previous study~\cite{zhong2015you}, we partition the whole Geolife dataset into one-week trajectories. We randomly sample 70\% for training (i.e., {\newCommentZheng{1,360 one-week trajectories}}) and the remaining for testing (i.e., {\newCommentZheng{584 one-week trajectories}}). 
For each trajectory, it is associated a ground truth label indicating the user's socioeconomic status.
%

{\newadd{To train the \texttt{DeepSEI}, we first pre-train the deep and recurrent networks separately and then jointly train them together. During the pre-training, it provides a warm-start for the two networks and boosts the convergence in the joint training. Specifically, we generate 50 episodes and use the deep or recurrent network separately for the classification task with cross entropy loss. During the joint training, we also generate 50 episodes and concatenate the outputs by the two networks as a long vector, which is further fed into a fully connected layer with the softmax function with the cross entropy loss.
We adopt the Adam stochastic gradient descent to optimize the network parameters. The training details are reported in Section~\ref{sec:result}.
}}
\section{EXPERIMENTS}
\label{sec:experiment}
\subsection{Experimental Setup}
\label{sec:setup}

\noindent \textbf{Datasets.}
The dataset used in this work contains three parts. 
The first part is the mobility data (i.e., GPS trajectories) generated by users.
The second part is POI data, which are used to capture users' activity features.
The third part is house price data, which is used to construct labels to reflect the socioeconomic statuses of users for evaluation. We collect the last two parts of data through web map services and web crawlers, respectively.

\noindent \textbf{1) Mobility Data.}
The mobility records correspond to a sequence of time-stamped locations sampled by the GPS device. For each record, it captures the location (i.e., latitude and longitude) of a user at a timestamp. We use the Geolife dataset~\cite{zheng2010geolife} for the mobility data. It contains 24,876,978 records in a period of five years, where the data is distributed in over 30 cities of China, and the majority part was generated in Beijing. 
{\newadd{We note that the dataset is publicly available without any personal information, avoiding possible privacy concerns.
}}

\noindent \textbf{2) POI Data.} 
We collect Point-of-Interest (POI) data through Amap Map API~\cite{api}, which contains 156,653 records in Beijing. For each POI record, it provides many attributes, including venue name, category, location with latitude and longitude information. We consider 11 major POI categories in this work, namely working, residence, food and drink, attractions, community, shopping, education,  hospitals, lodging, traffic, and recreation. We collect the POI data between 2011 and 2012, corresponding to the period of the mobility records on Geolife.

\noindent \textbf{3) House Price Data.} 
We collect the house price data by crawling online housing agents, i.e., Lianjia~\cite{lianjia}, which is the largest Chinese real-estate brokerage company that provides a comprehensive coverage of housing properties. We crawl 8,124 residential sale prices in Beijing. For each transaction, it records the residential name, sale price, floor size and residential address with latitude and longitude information we collected from Amap Map~\cite{api}. Here, the house prices correspond to the average prices with the unit of rmb/$m^2$. 
By following~\cite{xu2018human,ding2019estimating}, we use the house price data to indicate users' socioeconomic statuses. In particular, the range of collected average house price is from 10,588 to 113,224 in Beijing. {\nnCommentZheng{To study multiple-class classification, we take the binary classification for example. We define a binary label with the median threshold of the range (i.e., $\frac{10,588+113,224}{2}=61,891$), we set label 0 for the users, whose house prices are smaller than the threshold; label 1 otherwise. }}
Here, we find users' home locations by following the previous studies~\cite{phithakkitnukoon2012socio,xu2018human}, i.e., the home of a user is inferred as the location visited the most frequently during nighttime from 22:00 to 07:00.

\noindent \textbf{Tasks.
} 
We explore two tasks with the \texttt{DeepSEI} model. One is classification. We consider the number of classes from 2 to 5 by evenly partitioning the house price range to 2 - 5 equal intervals, as the corresponding socioeconomic class labels. The other is clustering. We collect the concatenated embeddings outputted by deep network and recurrent network, where the embeddings are learnt from the previous classification task, and thus they have incorporated socioeconomic context from the users.
We then explore $k$-means clustering on the concatenated embeddings, and vary the $k$ from 2 to 5. The class labels from 2 to 5 in the classification task are re-used as the ground truth for evaluating the clustering with $k$ from 2 to 5, respectively.


\begin{table*}[ht]
\centering
\caption{Effectiveness evaluation, where 2-5 denote the \# of classes or clusters.}
\vspace{-3mm}
\setlength{\tabcolsep}{5pt}
\begin{tabular}{|l|cccccccc|cccccccc|}
\hline
\multicolumn{1}{|c|}{Method}                  & \multicolumn{8}{c|}{Classification}                                                                                                                                                                            & \multicolumn{8}{c|}{Clustering}                                                                                                                                                                                                                                                                                                                  \\ \hline
\multicolumn{1}{|c|}{\multirow{2}{*}{Metric}} & \multicolumn{2}{c|}{2}                             & \multicolumn{2}{c|}{3}                             & \multicolumn{2}{c|}{4}                             & \multicolumn{2}{c|}{5}                          & \multicolumn{2}{c|}{2}                                                                  & \multicolumn{2}{c|}{3}                                                                  & \multicolumn{2}{c|}{4}                                                                  & \multicolumn{2}{c|}{5}                                             \\ \cline{2-17} 
\multicolumn{1}{|c|}{}                        & \multicolumn{1}{c|}{F1} & \multicolumn{1}{c|}{Acc} & \multicolumn{1}{c|}{F1} & \multicolumn{1}{c|}{Acc} & \multicolumn{1}{c|}{F1} & \multicolumn{1}{c|}{Acc} & \multicolumn{1}{c|}{F1} & Acc                   & \multicolumn{1}{c|}{ARI}                   & \multicolumn{1}{c|}{AMI}                   & \multicolumn{1}{c|}{ARI}                   & \multicolumn{1}{c|}{AMI}                   & \multicolumn{1}{c|}{ARI}                   & \multicolumn{1}{c|}{AMI}                   & \multicolumn{1}{c|}{ARI}                   & AMI                   \\ \hline
SES(RF)                                      & \multicolumn{1}{c|}{59.3}   & \multicolumn{1}{c|}{68.9}    & \multicolumn{1}{c|}{42.4}   & \multicolumn{1}{c|}{44.6}    & \multicolumn{1}{c|}{32.1}   & \multicolumn{1}{c|}{37.2}    & \multicolumn{1}{c|}{25.6}   & 30.3                      & \multicolumn{1}{c|}{\multirow{2}{*}{49.7}} & \multicolumn{1}{c|}{\multirow{2}{*}{50.2}} & \multicolumn{1}{c|}{\multirow{2}{*}{50.0}} & \multicolumn{1}{c|}{\multirow{2}{*}{50.2}} & \multicolumn{1}{c|}{\multirow{2}{*}{49.4}} & \multicolumn{1}{c|}{\multirow{2}{*}{50.7}} & \multicolumn{1}{c|}{\multirow{2}{*}{49.9}} & \multirow{2}{*}{50.3} \\ \cline{1-9}
SES(XGBoost)                                 & \multicolumn{1}{c|}{60.0}   & \multicolumn{1}{c|}{60.3}    & \multicolumn{1}{c|}{41.0}   & \multicolumn{1}{c|}{43.3}    & \multicolumn{1}{c|}{32.6}   & \multicolumn{1}{c|}{36.4}    & \multicolumn{1}{c|}{23.9}   &  27.8                     & \multicolumn{1}{c|}{}                      & \multicolumn{1}{c|}{}                      & \multicolumn{1}{c|}{}                      & \multicolumn{1}{c|}{}                      & \multicolumn{1}{c|}{}                      & \multicolumn{1}{c|}{}                      & \multicolumn{1}{c|}{}                      &                       \\ \hline
DIF(RF)                                      & \multicolumn{1}{c|}{69.1}   & \multicolumn{1}{c|}{69.3}    & \multicolumn{1}{c|}{57.1}   & \multicolumn{1}{c|}{59.6}    & \multicolumn{1}{c|}{50.3}   & \multicolumn{1}{c|}{53.0}    & \multicolumn{1}{c|}{47.1}   & 48.5                      & \multicolumn{1}{c|}{\multirow{2}{*}{60.3}} & \multicolumn{1}{c|}{\multirow{2}{*}{56.9}} & \multicolumn{1}{c|}{\multirow{2}{*}{57.8}}     & \multicolumn{1}{c|}{\multirow{2}{*}{55.9}}     & \multicolumn{1}{c|}{\multirow{2}{*}{54.0}}     & \multicolumn{1}{c|}{\multirow{2}{*}{52.6}}     & \multicolumn{1}{c|}{\multirow{2}{*}{50.7}}     & \multirow{2}{*}{50.4}     \\ \cline{1-9}
DIF(XGBoost)                                 & \multicolumn{1}{c|}{70.3}   & \multicolumn{1}{c|}{75.5}    &  \multicolumn{1}{c|}{61.6}    & \multicolumn{1}{c|}{63.5}   & \multicolumn{1}{c|}{52.6}    & \multicolumn{1}{c|}{55.7}   & \multicolumn{1}{c|}{42.6}    &   49.3                    & \multicolumn{1}{c|}{}                      & \multicolumn{1}{c|}{}                      & \multicolumn{1}{c|}{}                      & \multicolumn{1}{c|}{}                      & \multicolumn{1}{c|}{}                      & \multicolumn{1}{c|}{}                      & \multicolumn{1}{c|}{}                      &                       \\ \hline
L2P(RF)                                      & \multicolumn{1}{c|}{67.8}   & \multicolumn{1}{c|}{72.2}    & \multicolumn{1}{c|}{55.2}   & \multicolumn{1}{c|}{56.4}    & \multicolumn{1}{c|}{46.1}   & \multicolumn{1}{c|}{48.2}    & \multicolumn{1}{c|}{41.5}   &  44.3                     & \multicolumn{1}{c|}{\multirow{2}{*}{56.6}} & \multicolumn{1}{c|}{\multirow{2}{*}{55.5}} & \multicolumn{1}{c|}{\multirow{2}{*}{53.8}}     & \multicolumn{1}{c|}{\multirow{2}{*}{53.4}}     & \multicolumn{1}{c|}{\multirow{2}{*}{51.2}}     & \multicolumn{1}{c|}{\multirow{2}{*}{50.8}}     & \multicolumn{1}{c|}{\multirow{2}{*}{49.7}}     & \multirow{2}{*}{50.0}     \\ \cline{1-9}
L2P(XGBoost)                                 & \multicolumn{1}{l|}{68.2}   & \multicolumn{1}{l|}{67.8}    & \multicolumn{1}{l|}{58.4}   & \multicolumn{1}{l|}{59.3}    & \multicolumn{1}{l|}{47.7}   & \multicolumn{1}{l|}{49.0}    & \multicolumn{1}{l|}{40.2}   & 42.2  & \multicolumn{1}{c|}{}                      & \multicolumn{1}{c|}{}                      & \multicolumn{1}{c|}{}                      & \multicolumn{1}{c|}{}                      & \multicolumn{1}{c|}{}                      & \multicolumn{1}{c|}{}                      & \multicolumn{1}{c|}{}                      &                       \\ \hline
DeepSEI                                       & \multicolumn{1}{c|}{\textbf{86.1}}   & \multicolumn{1}{c|}{\textbf{90.2}}    & \multicolumn{1}{c|}{\textbf{80.2}}   & \multicolumn{1}{c|}{\textbf{80.4}}    & \multicolumn{1}{c|}{\textbf{63.9}}   & \multicolumn{1}{c|}{\textbf{64.2}}    & \multicolumn{1}{c|}{\textbf{53.3}}   &   \textbf{58.8}                    & \multicolumn{1}{c|}{\textbf{83.2}}                      & \multicolumn{1}{c|}{\textbf{77.5}}           
& \multicolumn{1}{c|}{\textbf{80.9}}                      & \multicolumn{1}{c|}{\textbf{71.9}}                      & \multicolumn{1}{c|}{\textbf{70.6}}                      & \multicolumn{1}{c|}{\textbf{70.3}}                      & \multicolumn{1}{c|}{\textbf{69.2}}                      &  \multicolumn{1}{c|}{\textbf{64.4}}                     \\ \hline
\end{tabular}
\vspace{-4mm}
\label{tab:deepmodels}
\end{table*}

\begin{table}[]
\centering
\caption{Ablation study for \texttt{DeepSEI}.}
\vspace{-3mm}
\begin{tabular}{l|cl|cl}
\hline
\multicolumn{1}{c|}{Method}     & \multicolumn{2}{c|}{Classification} & \multicolumn{2}{c}{Clustering} \\ \hline
\texttt{DeepSEI}                         & \multicolumn{2}{c|}{86.1}               & \multicolumn{2}{c}{83.2}           \\ \hline
w/o Deep Network                & \multicolumn{2}{c|}{73.7}               & \multicolumn{2}{c}{78.2}           \\
w/o Spatiality Diversity                  & \multicolumn{2}{c|}{78.8}               & \multicolumn{2}{c}{80.0}           \\
w/o Temporality Diversity                & \multicolumn{2}{c|}{82.5}               & \multicolumn{2}{c}{81.9}           \\
w/o Activity Diversity                    & \multicolumn{2}{c|}{82.8}               & \multicolumn{2}{c}{81.9}           \\ \hline
w/o Recurrent Network           & \multicolumn{2}{c|}{33.4}               & \multicolumn{2}{c}{60.4}           \\
w/o Spatial Feature          & \multicolumn{2}{c|}{75.7}               & \multicolumn{2}{c}{78.1}           \\
w/o Temporal Feature       & \multicolumn{2}{c|}{82.8}               & \multicolumn{2}{c}{80.5}           \\
w/o Semantic Feature & \multicolumn{2}{c|}{68.6}               & \multicolumn{2}{c}{77.9}           \\ \hline
\end{tabular}
\vspace{-5mm}
\label{tab:ablation}
\end{table}
\noindent \textbf{Baselines.} 
The following baselines are adapted.

\noindent $\bullet$ SES~\cite{xu2018human}. The study verified users' socioeconomic statuses can be reflected by their mobility patterns. 
Inspired by the study, we extract the mobility indicators including (1) radius of gyration, (2) number of activity locations, (3) activity entropy, (4) travel diversity, (5) K-radius of gyration, (6) unicity. 
Based on the features, we explore the following 10 classifiers for the classification task, including RBFSVM, LinearSVM, Logistic Regression, K-Nearest Neighbors (KNN), Decision Tree, Random Forest, Bayes, Adaboost, Gradient Boost and XGBoost. {\newadd{We grid search the best hyperparameters of the classifiers based on a development set.}}
Among them, we select two classifiers with the best effectiveness, namely SES (Random Forest) and SES (XGBoost). In addition, we concatenate those mobility indicators as a vector for the clustering task.

\noindent $\bullet$ DIF~\cite{wu2019inferring}. The study proposes a framework of inferring users’ demographics (e.g., gender, martial status or age) from their trajectories and geographical context. Other than the feature (1), (2), (3), (4), (6) studied in SES, it further incorporates (7) number of unique stay points, (8) centroid of stay points, (9) number of travels and (10) land use. Similarly, we explore the aforementioned 10 classifiers based on the features, and DIF (Random Forest) and DIF (XGBoost) dominate others in terms of the effectiveness. For clustering, we concatenate these features as the input.

\noindent $\bullet$ L2P~\cite{zhong2015you}. The study investigates users’ demographics and proposes a general location-to-profile (L2P) framework. L2P extracts the features from users' check-ins in terms of spatiality, temporality, and location knowledge (e.g., POI categories). It constructs a three-way tensor based on the extracted features, and adopts Tucker tensor decomposition to obtain a feature vector for each user. Based on the feature vectors, we still explore the aforementioned 10 classifiers, and L2P (Random Forest) and L2P (XGBoost) stand out. Those feature vectors can also be used for clustering.

\noindent \textbf{Evaluation Metrics.} 
Our paper involves two tasks: classification and clustering. For classification, we use the $F_1$-score and accuracy as the evaluation metric by following~\cite{zhong2015you}. 
For clustering, we use the metrics of Adjusted Rand Index (ARI) and Adjusted Mutual Information (AMI). They measure the correlation between the clustered result and the ground truth. Their values lie in the range of $[-1,1]$, and we normalize their values in $[0,1]$ for the ease of reading, where a higher ARI or AMI indicates a better result.

\noindent \textbf{Parameter Setting.}
Our \texttt{DeepSEI} model consists of deep network and recurrent network. For deep network, we embed the features of (1) spatiality diversity, (2) temporality diversity and (3) activity diversity into {\newCommentZheng{32-dimensional vectors}}, and concatenate them as a long vector with dimension {\newCommentZheng{$32*3=96$}} as the output. For recurrent network, we embed the features of (4) spatiality, (5) temporality and (6) activity into {\newCommentZheng{32-dimensional vectors}}, and feed the concatenation as a {\newCommentZheng{$96$-dimensional vector (i.e., $32*3=96$)}}, into a hierarchical LSTM. To implement the hierarchical LSTM, we use a low-level LSTM with {\newCommentZheng{64 hidden units}} and a high-level LSTM with {\newCommentZheng{64 hidden units}}. The hidden vector at the last step of the high-level LSTM is then fed into a fully connected layer with {\newCommentZheng{32 neurons}}. Thus, the output of recurrent network is a {\newCommentZheng{32-dimensional vector}}. Further, the outputs of deep network and recurrent network are concatenated as a {\newCommentZheng{$128$-dimensional vector (i.e., $96+32=128$)}}, which is fed into a fully connected layer with the softmax function as the activation function. We train the networks with Adam stochastic gradient descent and an initial learning rate of {\newCommentZheng{0.001.}} 

The default parameters of stay point radius $S_d$ and stay point duration $S_t$ are set to {\newCommentZheng{100m and 60 min}}, respectively. Here, we vary the parameter $S_d$ from {\newCommentZheng{100m to 300m}}, since the results are similar and we use the setting of {\newCommentZheng{$S_d=100$m.}} The setting of the parameter $S_t$ will be studied in experiments. For extracting the above features from (1) to (6), the following parameters are involved: the cell size for the feature (4), the spatiality granularity for the feature (1) and the temporality granularity and activity granularity for the feature (2,3). We use the settings of cell size, spatiality granularity, temporality granularity and activity granularity as {\newCommentZheng{200m, 100, 0.5 and 0.5}}, respectively. The results of their effects are shown in experiments.

\noindent \textbf{Evaluation Platform.}
We implement \texttt{DeepSEI} and other baselines in Python 3.6 and Tensorflow 2.3.0. 
The experiments are conducted on {\newCommentZheng{a desktop with Intel(R) Core(TM) i5-8265U CPU @ 1.60GHz 1.80 GHz and a 32 GB memory.}}
%
%
The codes can be downloaded via the link~\footnote{\url{https://github.com/zhengwang125/DeepSEI}}.

\subsection{Experimental Results}
\label{sec:result}

\noindent \textbf{(1) Effectiveness evaluation (comparison with different classifiers).}
We compare the \texttt{DeepSEI} model with the baselines. In Table~\ref{tab:deepmodels}, we report their effectiveness in terms of $F_1$-score(\%) and accuracy(\%) for classification and ARI(\%) and AMI(\%) for clustering. Overall, our \texttt{DeepSEI} model consistently outperforms the baselines, e.g., in binary classification and clustering, it outperforms the best baseline (i.e., DIF) by 22.5\% and 37.9\%, respectively. 
The reasons are mainly two-fold: 1) the \texttt{DeepSEI} model is with more comprehensive features to infer the users' socioeconomic statuses from three aspects, i.e., spatiality, temporality and activity; 2) the two networks that are incorporated by DeepSEI can capture the features effectively as they capture the features at both the coarse and detailed levels.
%


\noindent \textbf{(2) Ablation study.}
We conduct an ablation study to evaluate the effect of each network (i.e., deep network or recurrent network) and features in the \texttt{DeepSEI} model, and the comparing results are reported in Table~\ref{tab:ablation}. Overall, we can see all these networks and features contribute to the final result.
For the recurrent Network, w/o Recurrent Network corresponding to the case that only Deep Network is kept, the result performs the worst with $F_1$-score of 33.4\% and ARI of 60.4\%. This is because it captures a sequence of users' daily activities, which is essential to infer users' socioeconomic statuses.
For the deep network, we observe the spatiality diversity is with the most effect, e.g., when the spatiality is removed, the $F_1$-score is 78.8, which drops by 9.3\%. This is because users' socioeconomic statuses are highly linked to the range of their activity territory, which has been verified in previous studies~\cite{xu2018human}.

\begin{table}[t]
\centering
\caption{Impacts of stay point duration (mins) for \texttt{DeepSEI}.}
\vspace{-3mm}
\setlength{\tabcolsep}{6pt}
\begin{tabular}{lccccc}
\hline
Parameter & 30 & 60 & 90 & 120 & 150 \\ \hline
Classification   &74.3     &  78.9   &  \textbf{86.1}  &  82.6   &  81.9   \\
Clustering   & 80.9    &  81.9   &  \textbf{83.2}  &   81.9  &  81.8  \\ \hline
\label{tab:duration}
\end{tabular}
\vspace*{-4mm}
\end{table}

\begin{table}[t]
\centering
\caption{Impacts of cell size (m) for \texttt{DeepSEI}.}
\vspace{-3mm}
\setlength{\tabcolsep}{6.5pt}
\begin{tabular}{lccccc}
\hline
Parameter & 100 & 200 & 300 & 400 & 500 \\ \hline
Classification   &82.3     &  \textbf{86.1}   &  83.6  &  81.5   &   79.8  \\ 
Clustering   &82.0     &  \textbf{83.2}   &  81.2  &  80.5   & 79.4    \\ \hline
\end{tabular}
\label{tab:cellsize}
\vspace*{-4mm}
\end{table}

\begin{table}[t]
\centering
\caption{Impacts of spatiality granularity for \texttt{DeepSEI}.}
\vspace{-3mm}
\setlength{\tabcolsep}{7.9pt}
\begin{tabular}{lccccc}
\hline
Parameter & 100 & 200 & 300 & 400 & 500 \\ \hline
Classification & 84.1 & 85.9 & \textbf{86.1} & 85.6 & 83.8 \\ 
Clustering & 82.1 & 81.9 & \textbf{83.2} & 82.7 & 82.0  \\ \hline
\end{tabular}
\label{tab:spatiality}
\vspace*{-4mm}
\end{table}

\noindent \textbf{(3) Parameter study (varying stay point duration $S_t$).}
The stay point duration parameter $S_t$ controls the time threshold of the stay point detection algorithm, where the points in a trajectory will be merged as one stay point if the duration of the first point (called anchor point) and the last point in the trajectory is within $S_t$. We vary the $S_t$ from 30 minutes to 150 minutes, and the results are reported in Table~\ref{tab:ablation}. As expected, with the increase of $S_t$, less stay points are detected, which corresponds to less training and testing instances are generated.
We observe that our model performs the best when $S_t$ is set to 90 minutes with the $F_1$-score of 86.1\% and ARI of 83.2\%. This is because with a small $S_t$, the detected stay points cannot accurately reflect the users' activities, it falsely takes some trivial behaviors such as ``walking'' as the users' activities, which prevents the model to learn useful information. With a large $S_t$, many stay points cannot be detected, which causes the model performance degrades as many features that are associated with the stay points are missing.

\begin{table}[t]
\centering
\caption{Impacts of temporality and activity granularity for \texttt{DeepSEI}.}
\vspace{-3mm}
\setlength{\tabcolsep}{7.5pt}
\begin{tabular}{lccccc}
\hline
Parameter & 0.1 & 0.3 & 0.5 & 0.7 & 0.9 \\ \hline
Classification   &72.3     &  78.6   &  \textbf{86.1}  &  83.8   &  81.6   \\ 
Clustering   &80.4     & 81.5    & \textbf{83.2}   &  81.8   & 80.8    \\\hline
\label{tab:embtokens}
\end{tabular}
\vspace*{-6mm}
\end{table}

\noindent \textbf{(4) Parameter study (varying cell size).}
In Table~\ref{tab:cellsize}, we study the effect of cell size by varying the size from 100 meters to 500 meters. Here, the effect of cell size is mainly in two aspects: 1) for spatiality, each grids contains a set of stay points; with a large size, more stay points will be indexed to the same grid, which results many stay points share the same mobility embedding; 2) for activity, we infer the activity of a stay point using the POIs in 8 neighbouring grids of the grid, where the stay point is located; then, a large cell size, corresponds to a large grid, which takes more POIs into consideration. Based on the results, we observe that the cell size 200 meters fits our model best. With the smallest cell size (100 meters), the model degrades. This is because a small cell size generates more tokens, which makes the model difficult to train. On the other hand, a large cell size causes a lower resolution of the stay points, and overlooks the differences of mobility features. In addition, as more POIs are considered in a large grid, those POIs may generate noise to interfere with POI inference and this is in line with our intuition.

\noindent \textbf{(5) Parameter study (varying spatiality diversity granularity).} 
We study the effect of spatiality granularity in Table~\ref{tab:spatiality}. We vary the granularity from 100 to 500, and report the effectiveness in terms of classification and clustering. The parameter captures the resolution of the spatial range of users' daily activities. A smaller value provides a higher resolution but incurs more tokens, which affects the model training. With a larger value, the capability of model to distinguish different spatial diversities will degrade. We set it to 300, which leads to the best effectiveness.

\noindent \textbf{(6) Parameter study (varying temporality and activity diversity granularity).}
We study the effect of diversity granularity for two entropy-based indicators, i.e., temporality diversity and activity diversity. We vary the granularity parameter from 0.1 to 0.9, and the results are reported in Table~\ref{tab:embtokens}. Temporality diversity and activity diversity are two features with continuous numerical values, dividing their ranges with a small granularity will generate too many tokens, and causes the model difficult to learn the features; with a large granularity, the capability of the model to identify the diversity from users' mobility patterns will degrade, which is as expected. overall, a moderate setting with the value of 0.5 provides the best effectiveness.

\begin{figure}[t]
\centering
\begin{tabular}{c c}
   \begin{minipage}{0.47\linewidth}
    \includegraphics[width=\linewidth]{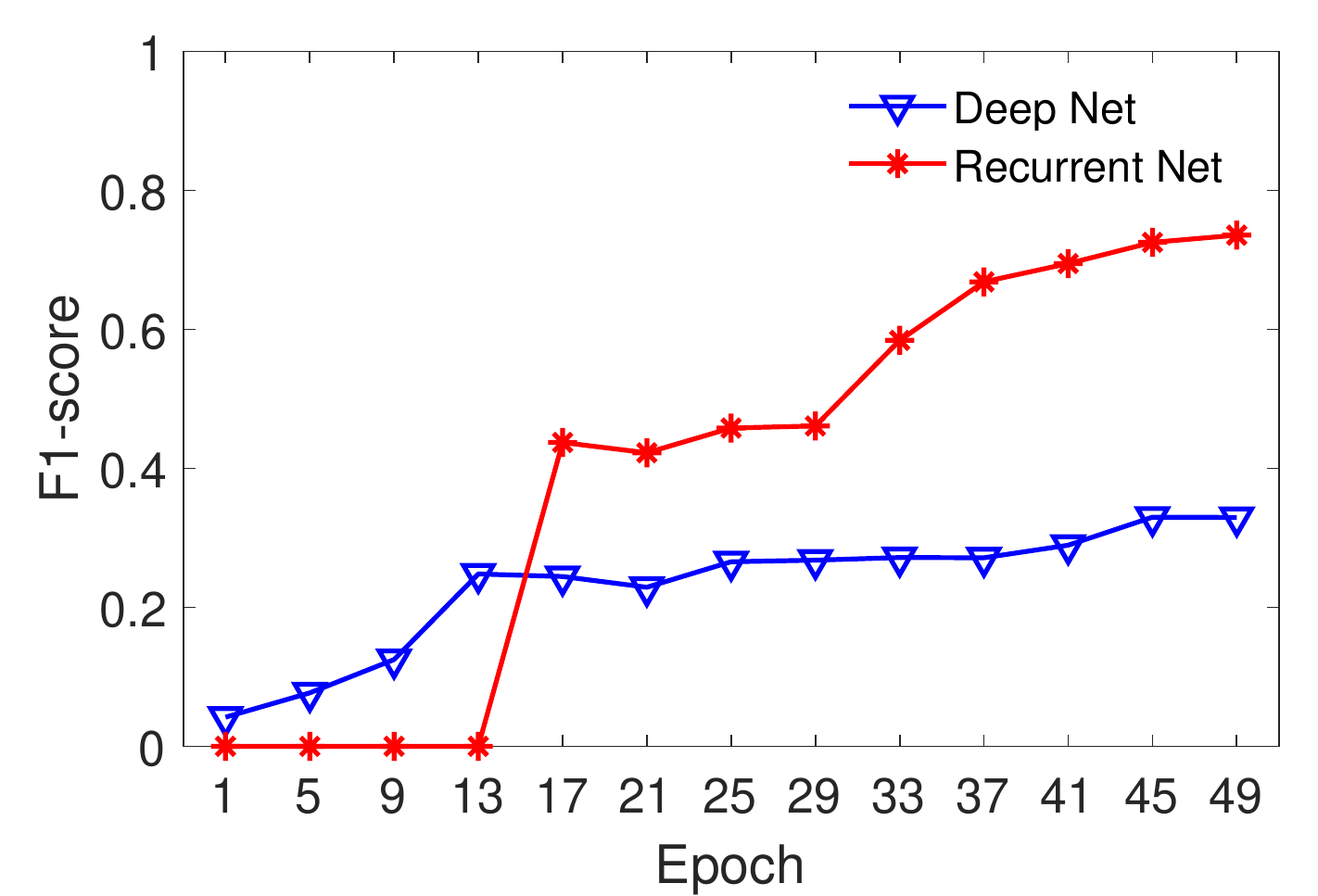}
    \end{minipage}
    &
    \begin{minipage}{0.47\linewidth}
    \includegraphics[width=\linewidth]{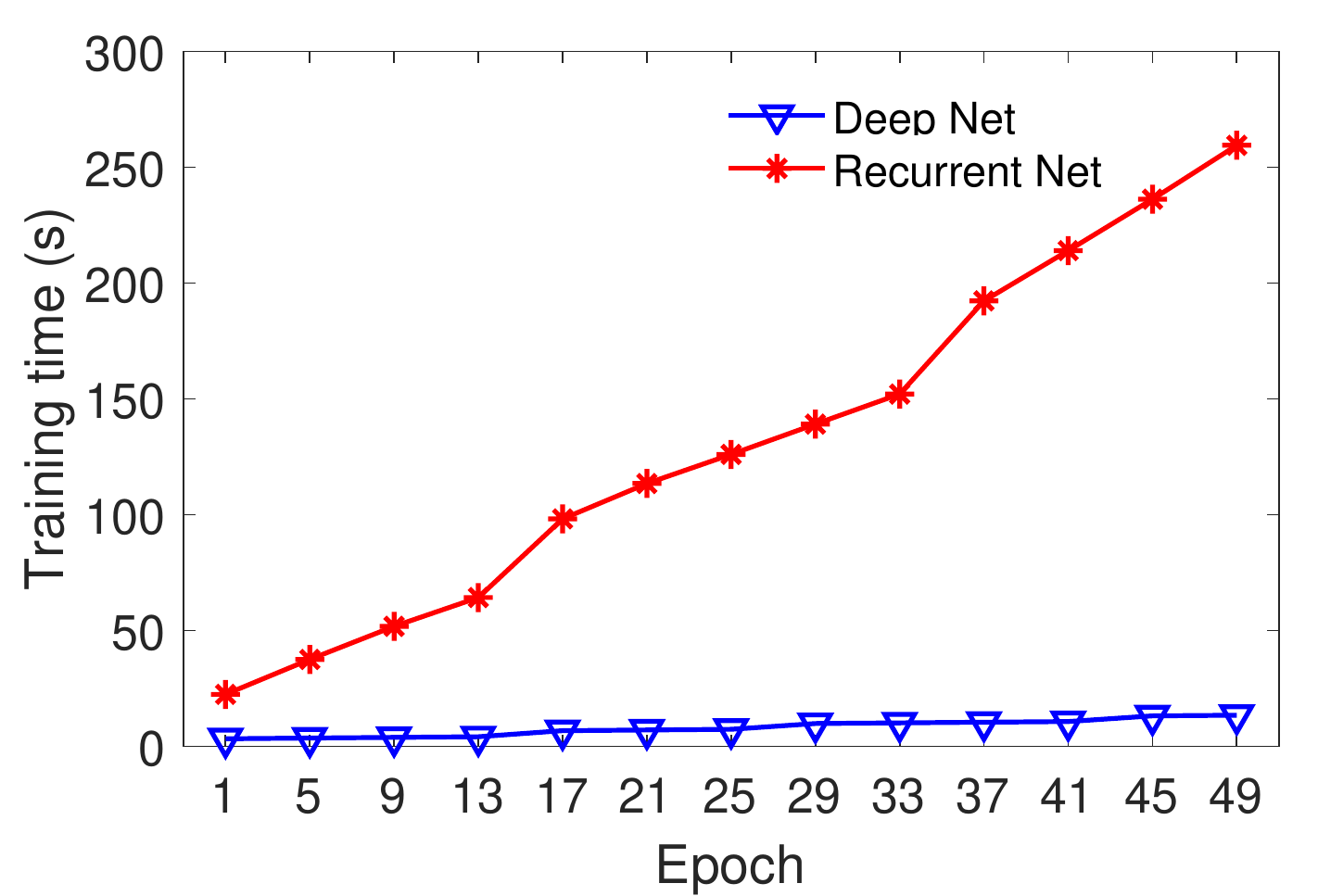}
    \end{minipage}
    \\
    (a) Pre-training ($F_1$-score)
    &
    (b) Pre-training (Time cost)
    \\
      \begin{minipage}{0.47\linewidth}
    \includegraphics[width=\linewidth]{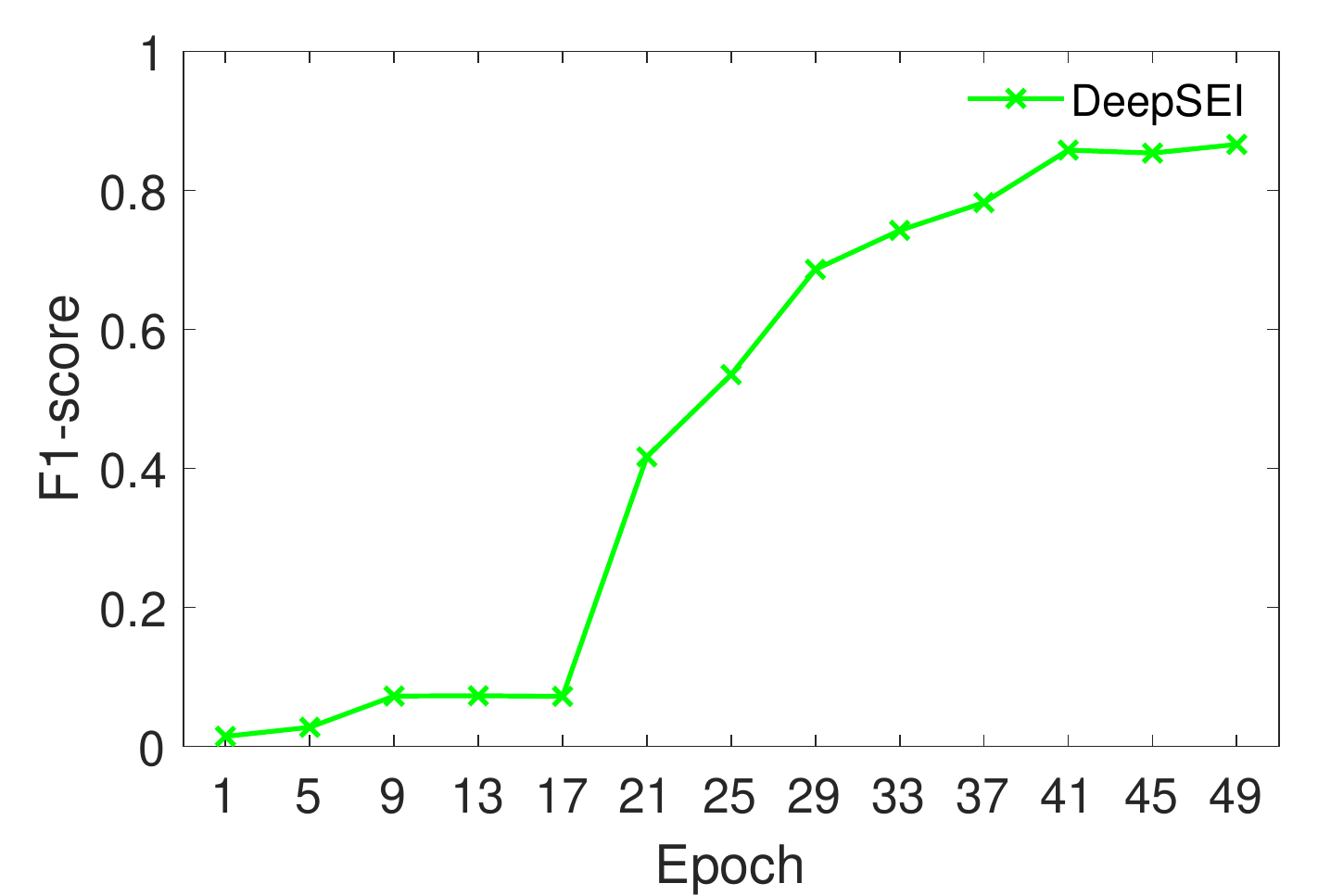}
    \end{minipage}
    &
    \begin{minipage}{0.47\linewidth}
    \includegraphics[width=\linewidth]{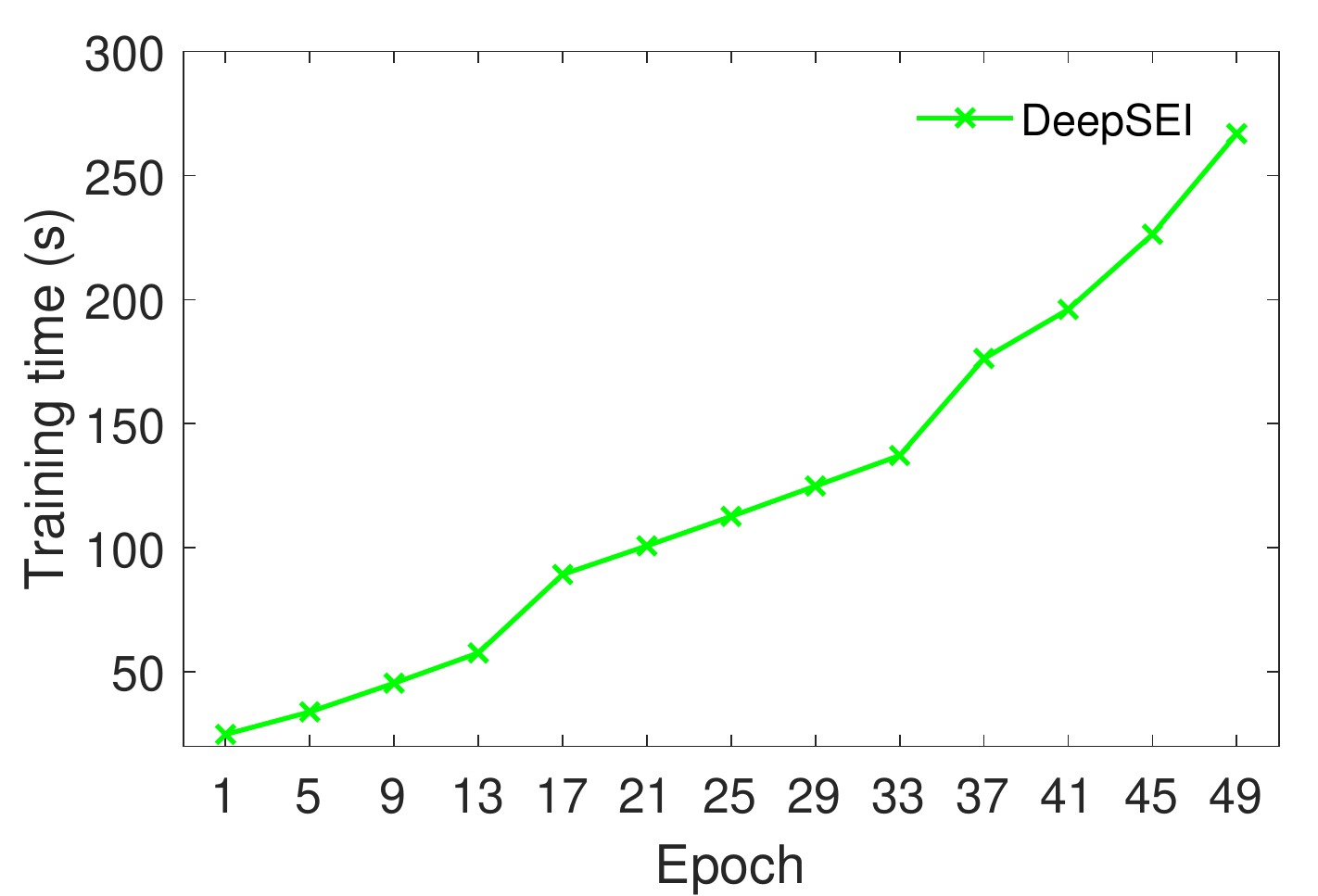}
    \end{minipage}
    \\
    (c) Training ($F_1$-score)
    &
    (d) Training (Time cost)
\end{tabular}
\vspace*{-2mm}
\caption{Training cost on Geolife.}
\label{fig:train}
\vspace*{-3mm}
\end{figure}
\noindent \textbf{(7) Training time.} In Figure~\ref{fig:train}, we report the times and the corresponding effectiveness with the default setup in Section~\ref{sec:setup}. We generate 50 epochs for both pre-training and training.
We observe that the effectiveness improves with the number of epochs and the corresponding training time increases almost linearly. In pre-training, the recurrent network takes more time because it has a more complex network architecture (i.e., hierarchical LSTM). In training, the \texttt{DeepSEI} model incorporates the two networks and obtains a further improvement after 32 epochs. We observe that the \texttt{DeepSEI} model converges after 41 epochs, and we use the trained model for other experiments.

\begin{figure*}[ht]
	\centering
	\begin{tabular}{c c c c}
		\begin{minipage}{0.24\linewidth}
			\includegraphics[width=\linewidth]{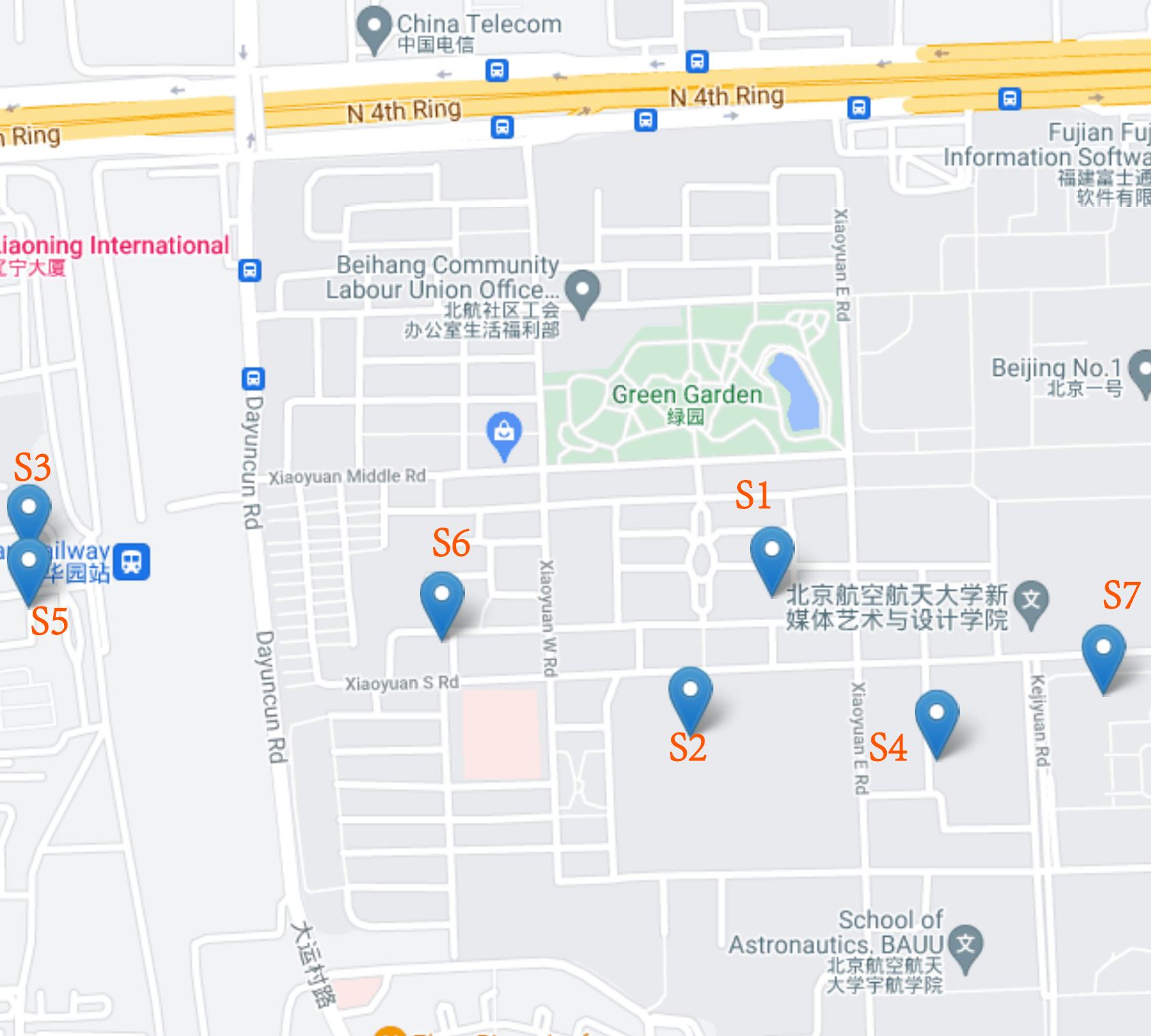}
		\end{minipage}
		&
		\begin{minipage}{0.23\linewidth}
			\includegraphics[width=\linewidth]{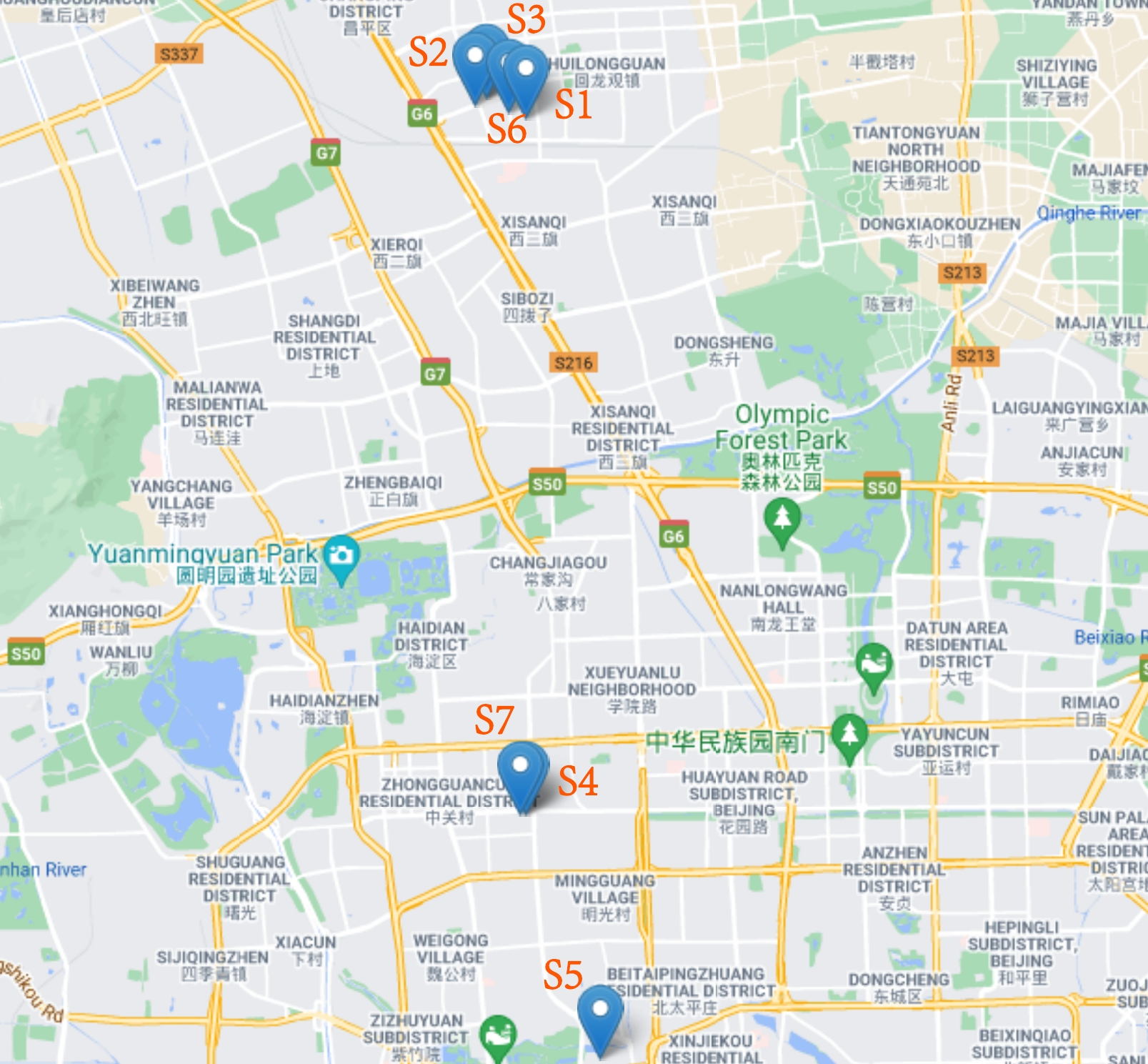}
		\end{minipage}
		&
		\begin{minipage}{0.24\linewidth}
			\includegraphics[width=\linewidth]{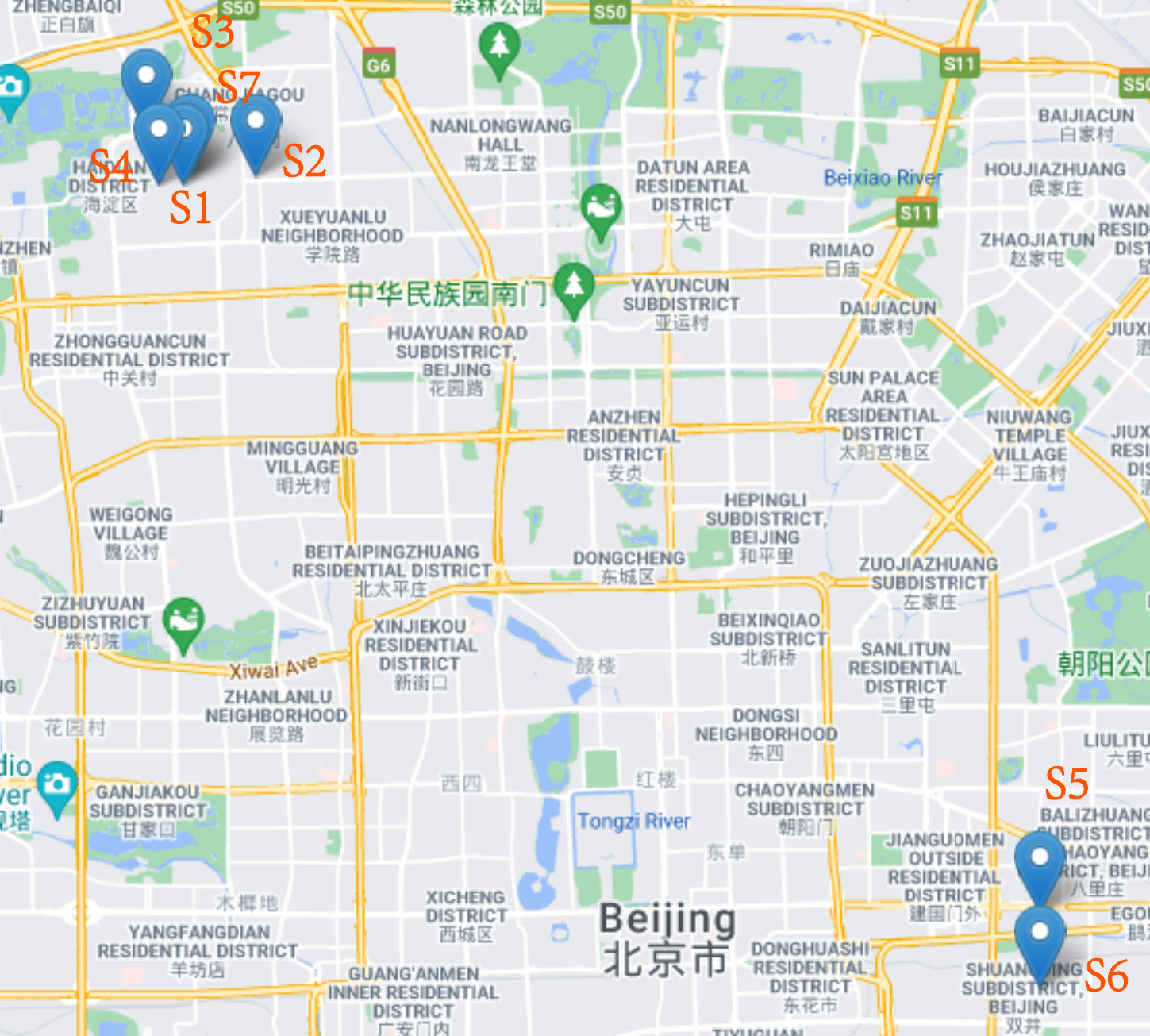}
		\end{minipage}
		&
		\begin{minipage}{0.23\linewidth}
			\includegraphics[width=\linewidth]{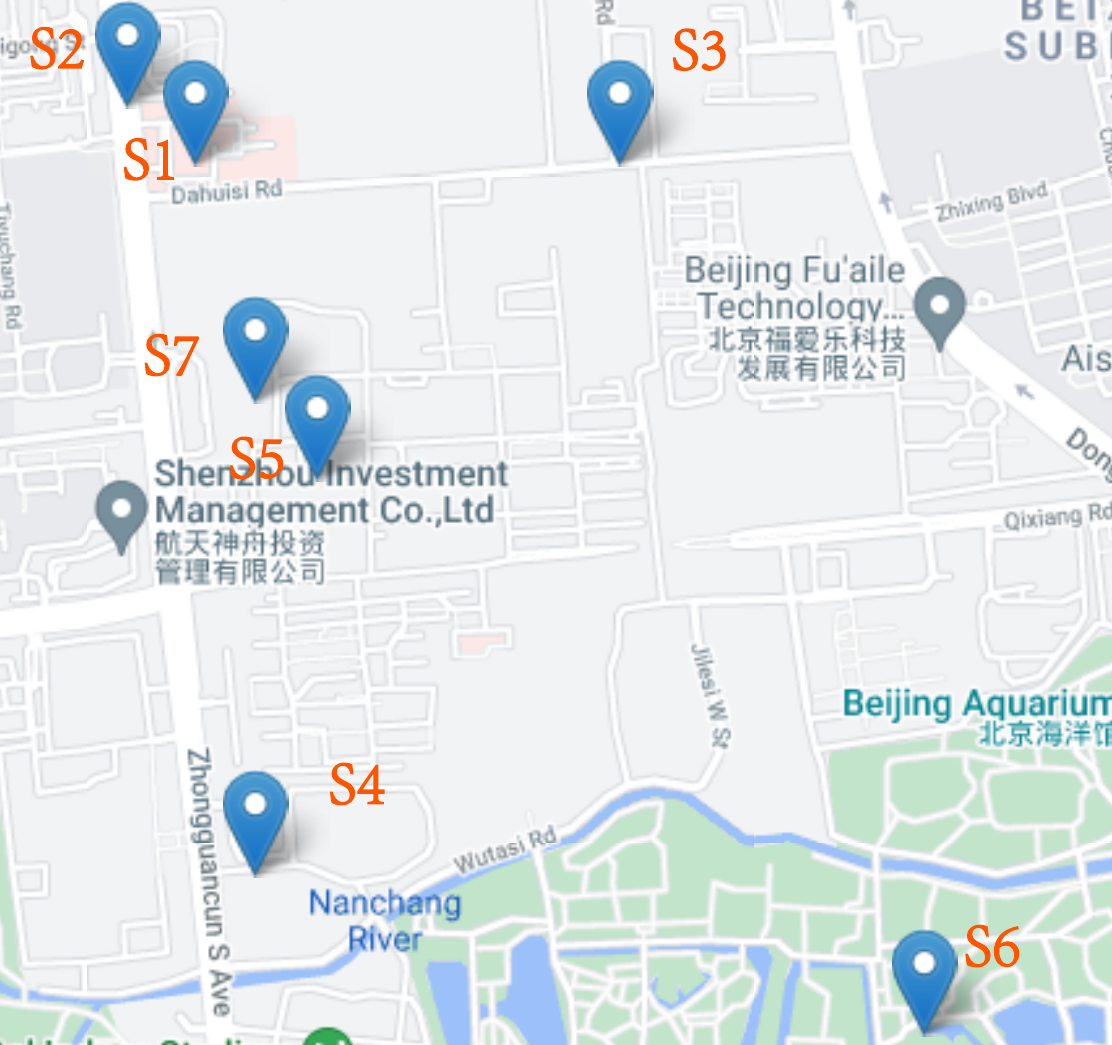}
		\end{minipage}
		\\
		(a) User 1 (richer)
		&
		(b) User 2 (richer)
		&
		(c) User 3 (poorer)
		&
		(d) User 4 (poorer)
	\end{tabular}
	\vspace*{-3mm}
	\caption{Illustration of stay points for four users with different socioeconomic statuses.}
	\label{fig:case}
	\vspace*{-3mm}
\end{figure*}

\begin{table*}
\centering
\caption{Case study, DS, DT and DA denote the features captured via the deep network for spatiality diversity, temporality diversity and activity diversity; RT and RS denote that via the recurrent network for temporal (time bin) and semantic features.}
\vspace*{-3mm}
\setlength{\tabcolsep}{11pt}
\begin{tabular}{|c|cc|cc|cc|cc|}
\hline
Case          & \multicolumn{2}{c|}{User 1 (richer)} & \multicolumn{2}{c|}{User 2 (richer)} & \multicolumn{2}{c|}{User 3 (poorer)} & \multicolumn{2}{c|}{User 4 (poorer)} \\ \hline
DS, DT and DA & \multicolumn{2}{c|}{9.62, 1.30 and 2.16}                 & \multicolumn{2}{c|}{5.52, 3.14 and 2.20}                 & \multicolumn{2}{c|}{14.29, 4.45 and 2.77}                 & \multicolumn{2}{c|}{10.38, 4.89 and 5.12}                 \\ \hline
Stay points   & \multicolumn{1}{c|}{RT}      & RS     & \multicolumn{1}{c|}{RT}      & RS     & \multicolumn{1}{c|}{RT}      & RS     & \multicolumn{1}{c|}{RT}      & RS     \\ \hline
$s_1$       & \multicolumn{1}{c|}{18}        & residence       &  \multicolumn{1}{c|}{20}        & residence      & \multicolumn{1}{c|}{9}        & working        & \multicolumn{1}{c|}{32}        & hospital       \\ \hline
$s_2$       & \multicolumn{1}{c|}{7}        &  recreation      & \multicolumn{1}{c|}{12}        & food and drink      & \multicolumn{1}{c|}{10}        & traffic        & \multicolumn{1}{c|}{35}        & traffic        \\ \hline
$s_3$       & \multicolumn{1}{c|}{9}        & education       & \multicolumn{1}{c|}{12}        & traffic      & \multicolumn{1}{c|}{12}        & food and drink        & \multicolumn{1}{c|}{36}        & food and drink       \\ \hline
$s_4$       & \multicolumn{1}{c|}{10}        &  working      & \multicolumn{1}{c|}{14}        &    working & \multicolumn{1}{c|}{13}        & working        & \multicolumn{1}{c|}{40}        & community       \\ \hline
$s_5$       & \multicolumn{1}{c|}{13}        &  education      & \multicolumn{1}{c|}{19}        & lodging      & \multicolumn{1}{c|}{18}        & residence        & \multicolumn{1}{c|}{41}        &    residence    \\ \hline
$s_6$       & \multicolumn{1}{c|}{17}        &  residence      & \multicolumn{1}{c|}{37}        & residence     & \multicolumn{1}{c|}{7}        &  traffic        & \multicolumn{1}{c|}{34}        & attractions       \\ \hline
$s_7$       & \multicolumn{1}{c|}{9}        & working       & \multicolumn{1}{c|}{8}        &   working  & \multicolumn{1}{c|}{8}        &   working        & \multicolumn{1}{c|}{43}        & residence       \\ \hline
\end{tabular}
\label{tab:case}
\vspace*{-2mm}
\end{table*}
\noindent \textbf{(8) Case study.} We conduct a case study. We select four cases for the study, where User 1 and 2 are identified as the richer users in the same class 1, and User 3 and 4 are identified as the poorer users in another class 2.
In Figure~\ref{fig:case}, we visualize the locations of their stay points on the map. In Table~\ref{tab:case}, we list the features captured by the deep network and recurrent network. We observe the following insights that may explain the relationship between their mobility patterns and socioeconomic statuses.
\\
\emph{\underline{Insight 1}: Richer users tend to travel shorter.} In Table~\ref{tab:case}, we observe the richer users (User 1 and User 2) are generally with the smaller spatiality diversities (e.g., 9.62 and 5.52) than poorer users. This insight is in line with the intuition from the previous study~\cite{xu2018human}, and the reason could be that rich people are busy with work and have limited time for travelling.
\\
\emph{\underline{Insight 2}: Richer users are generally with lower temporality/activity diversity of daily activities.} Temporality/Activity diversity is an entropy-based feature to reveal the regularity of users' daily activities. 
In Table~\ref{tab:case}, we observe the regularity of User 1 and User 2 are high, corresponding to the smaller values. For example, in Figure~\ref{fig:case}, User 1  mainly commutes between home (``residence'' POI) and office (``working'' POI) regularly, and he/she is with the least temporality diversity 1.30. In contrast, the User 4 is irregular, e.g., he/she visits many places instead of staying somewhere and working. 
\\
\emph{\underline{Insight 3}: Richer users are with secure jobs.} We infer the users' employment statuses based on the data extracted from their stay points. We infer that User 1 is with a steady job since he/she works (at 09:00 am - 05:00 pm) and stay homes (at 05:00 pm - 07:00 am) regularly. In this situation, he/she has a stable source of income (e.g., we infer that he/she may be a faculty at an university based on the stay points on the map), and the status is reflected on his/her house price data accordingly.
%

\section{CONCLUSION}
\label{sec:conclusion}

In this paper, 
we study user socioeconomic status inference, and propose a novel socioeconomic-aware deep model called \texttt{DeepSEI} for the task. \texttt{DeepSEI} incorporates two neural networks, i.e., deep network and recurrent network. 
%
We extract features from three aspects of users' mobility records, namely spatiality, temporality and activity, at the coarse and detailed levels, and feed them to the two networks.
We conduct the experiments on Geolife dataset, POI dataset and house price dataset, and the results show that the \texttt{DeepSEI} model achieves a noticeable improvement over the baselines. 
In the future, we plan to explore more mining and learning tasks based on users' mobility records, such as anomaly detection.

\smallskip
\noindent\textbf{Acknowledgments:}
This research/project is supported by the National Research Foundation, Singapore under its AI Singapore Programme (AISG Award No: AISG-PhD/2021-08-024[T]).
This research is also supported by the Ministry of Education, Singapore, under its Academic Research Fund (Tier 2 Awards MOE-T2EP20220-0011 and MOE-T2EP20221-0013). 
This research is also supported in part by the China-Singapore International Joint Research Institute (CSIJRI), Guangzhou, China (Award No. 206-A021002).
Any opinions, findings and conclusions or recommendations expressed in this material are those of the author(s) and do not reflect the views of National Research Foundation, Singapore and Ministry of Education, Singapore. This project is partially supported by HKU-SCF FinTech Academy and also Shenzhen Science and Technology Innovation Committee (SZ-HK-Macau Technology Research Programme, \#SGDX20210823103537030).

\bibliography{ref3}

\begin{thebibliography}{10}
\providecommand{\url}[1]{#1}
\csname url@samestyle\endcsname
\providecommand{\newblock}{\relax}
\providecommand{\bibinfo}[2]{#2}
\providecommand{\BIBentrySTDinterwordspacing}{\spaceskip=0pt\relax}
\providecommand{\BIBentryALTinterwordstretchfactor}{4}
\providecommand{\BIBentryALTinterwordspacing}{\spaceskip=\fontdimen2\font plus
\BIBentryALTinterwordstretchfactor\fontdimen3\font minus
  \fontdimen4\font\relax}
\providecommand{\BIBforeignlanguage}[2]{{%
\expandafter\ifx\csname l@#1\endcsname\relax
\typeout{** WARNING: IEEEtran.bst: No hyphenation pattern has been}%
\typeout{** loaded for the language `#1'. Using the pattern for}%
\typeout{** the default language instead.}%
\else
\language=\csname l@#1\endcsname
\fi
#2}}
\providecommand{\BIBdecl}{\relax}
\BIBdecl

\bibitem{zhong2015you}
Y.~Zhong, N.~J. Yuan, W.~Zhong, F.~Zhang, and X.~Xie, ``You are where you go:
  Inferring demographic attributes from location check-ins,'' in \emph{WSDM},
  2015, pp. 295--304.

\bibitem{riederer2020location}
C.~Riederer, \emph{Location Data: Perils, Profits, Promise}.\hskip 1em plus
  0.5em minus 0.4em\relax Columbia University, 2020.

\bibitem{zhang2019deep}
Y.~Zhang and T.~Cheng, ``A deep learning approach to infer employment status of
  passengers by using smart card data,'' \emph{TITS}, vol.~21, no.~2, pp.
  617--629, 2019.

\bibitem{ding2019estimating}
S.~Ding, H.~Huang, T.~Zhao, and X.~Fu, ``Estimating socioeconomic status via
  temporal-spatial mobility analysis-a case study of smart card data,'' in
  \emph{ICCCN}.\hskip 1em plus 0.5em minus 0.4em\relax IEEE, 2019, pp. 1--9.

\bibitem{wu2019inferring}
L.~Wu, L.~Yang, Z.~Huang, Y.~Wang, Y.~Chai, X.~Peng, and Y.~Liu, ``Inferring
  demographics from human trajectories and geographical context,''
  \emph{Computers, Environment and Urban Systems}, vol.~77, p. 101368, 2019.

\bibitem{xu2018human}
Y.~Xu, A.~Belyi, I.~Bojic, and C.~Ratti, ``Human mobility and socioeconomic
  status: Analysis of singapore and boston,'' \emph{Computers, Environment and
  Urban Systems}, vol.~72, pp. 51--67, 2018.

\bibitem{hanson1981travel}
S.~Hanson and P.~Hanson, ``The travel-activity patterns of urban residents:
  dimensions and relationships to sociodemographic characteristics,''
  \emph{Economic geography}, pp. 332--347, 1981.

\bibitem{kwan1999gender}
M.-P. Kwan, ``Gender, the home-work link, and space-time patterns of
  nonemployment activities,'' \emph{Economic geography}, vol.~75, no.~4, pp.
  370--394, 1999.

\bibitem{limtanakool2006influence}
N.~Limtanakool, M.~Dijst, and T.~Schwanen, ``The influence of socioeconomic
  characteristics, land use and travel time considerations on mode choice for
  medium-and longer-distance trips,'' \emph{Journal of transport geography},
  vol.~14, no.~5, pp. 327--341, 2006.

\bibitem{xu2015understanding}
Y.~Xu, S.-L. Shaw, Z.~Zhao, L.~Yin, Z.~Fang, and Q.~Li, ``Understanding
  aggregate human mobility patterns using passive mobile phone location data: a
  home-based approach,'' \emph{Transportation}, vol.~42, no.~4, pp. 625--646,
  2015.

\bibitem{blumenstock2015predicting}
J.~Blumenstock, G.~Cadamuro, and R.~On, ``Predicting poverty and wealth from
  mobile phone metadata,'' \emph{Science}, vol. 350, no. 6264, pp. 1073--1076,
  2015.

\bibitem{huang2016activity}
Q.~Huang and D.~W. Wong, ``Activity patterns, socioeconomic status and urban
  spatial structure: what can social media data tell us?'' \emph{IJGIS},
  vol.~30, no.~9, pp. 1873--1898, 2016.

\bibitem{kelly2013uncovering}
D.~Kelly, B.~Smyth, and B.~Caulfield, ``Uncovering measurements of social and
  demographic behavior from smartphone location data,'' \emph{THMS}, vol.~43,
  no.~2, pp. 188--198, 2013.

\bibitem{monreale2009wherenext}
A.~Monreale, F.~Pinelli, R.~Trasarti, and F.~Giannotti, ``Wherenext: a location
  predictor on trajectory pattern mining,'' in \emph{SIGKDD}, 2009, pp.
  637--646.

\bibitem{zhang2014splitter}
C.~Zhang, J.~Han, L.~Shou, J.~Lu, and T.~La~Porta, ``Splitter: Mining
  fine-grained sequential patterns in semantic trajectories,'' \emph{PVLDB},
  vol.~7, no.~9, pp. 769--780, 2014.

\bibitem{chen2020context}
Y.~Chen, C.~Long, G.~Cong, and C.~Li, ``Context-aware deep model for joint
  mobility and time prediction,'' in \emph{WSDM}, 2020, pp. 106--114.

\bibitem{ju2020interaction}
C.~Ju, Z.~Wang, C.~Long, X.~Zhang, and D.~E. Chang, ``Interaction-aware kalman
  neural networks for trajectory prediction,'' in \emph{IV}.\hskip 1em plus
  0.5em minus 0.4em\relax IEEE, 2020, pp. 1793--1800.

\bibitem{feng2018deepmove}
J.~Feng, Y.~Li, C.~Zhang, F.~Sun, F.~Meng, A.~Guo, and D.~Jin, ``Deepmove:
  Predicting human mobility with attentional recurrent networks,'' in
  \emph{WWW}, 2018, pp. 1459--1468.

\bibitem{xiao2017modeling}
S.~Xiao, J.~Yan, X.~Yang, H.~Zha, and S.~Chu, ``Modeling the intensity function
  of point process via recurrent neural networks,'' in \emph{AAAI}, vol.~31,
  no.~1, 2017.

\bibitem{du2016recurrent}
N.~Du, H.~Dai, R.~Trivedi, U.~Upadhyay, M.~Gomez-Rodriguez, and L.~Song,
  ``Recurrent marked temporal point processes: Embedding event history to
  vector,'' in \emph{SIGKDD}, 2016, pp. 1555--1564.

\bibitem{mohamed2016clustering}
K.~Mohamed, E.~C{\^o}me, L.~Oukhellou, and M.~Verleysen, ``Clustering smart
  card data for urban mobility analysis,'' \emph{TITS}, vol.~18, no.~3, pp.
  712--728, 2016.

\bibitem{zheng2015trajectory}
Y.~Zheng, ``Trajectory data mining: an overview,'' \emph{TIST}, vol.~6, no.~3,
  pp. 1--41, 2015.

\bibitem{li2008mining}
Q.~Li, Y.~Zheng, X.~Xie, Y.~Chen, W.~Liu, and W.-Y. Ma, ``Mining user
  similarity based on location history,'' in \emph{SIGSPATIAL}, 2008, pp.
  1--10.

\bibitem{wu2016did}
F.~Wu and Z.~Li, ``Where did you go: Personalized annotation of mobility
  records,'' in \emph{CIKM}, 2016, pp. 589--598.

\bibitem{yan2013semantic}
Z.~Yan, D.~Chakraborty, C.~Parent, S.~Spaccapietra, and K.~Aberer, ``Semantic
  trajectories: Mobility data computation and annotation,'' \emph{TIST},
  vol.~4, no.~3, pp. 1--38, 2013.

\bibitem{yan2011semitri}
------, ``Semitri: a framework for semantic annotation of heterogeneous
  trajectories,'' in \emph{EDBT}, 2011, pp. 259--270.

\bibitem{scheiner2014gendered}
J.~Scheiner, ``The gendered complexity of daily life: effects of life-course
  events on changes in activity entropy and tour complexity over time,''
  \emph{Travel Behaviour and Society}, vol.~1, no.~3, pp. 91--105, 2014.

\bibitem{pappalardo2015using}
L.~Pappalardo, D.~Pedreschi, Z.~Smoreda, and F.~Giannotti, ``Using big data to
  study the link between human mobility and socio-economic development,'' in
  \emph{BigData}.\hskip 1em plus 0.5em minus 0.4em\relax IEEE, 2015, pp.
  871--878.

\bibitem{mikolov2013efficient}
T.~Mikolov, K.~Chen, G.~Corrado, and J.~Dean, ``Efficient estimation of word
  representations in vector space,'' \emph{arXiv}, 2013.

\bibitem{hochreiter1997long}
S.~Hochreiter and J.~Schmidhuber, ``Long short-term memory,'' \emph{Neural
  computation}, vol.~9, no.~8, pp. 1735--1780, 1997.

\bibitem{zheng2010geolife}
Y.~Zheng, X.~Xie, W.-Y. Ma \emph{et~al.}, ``Geolife: A collaborative social
  networking service among user, location and trajectory.'' \emph{IEEE Data
  Eng. Bull.}, vol.~33, no.~2, pp. 32--39, 2010.

\bibitem{api}
{Amap Map Service}, \url{https://lbs.amap.com/api/webservice/summary/}.

\bibitem{lianjia}
{Homelink}, \url{https://www.lianjia.com}.

\bibitem{phithakkitnukoon2012socio}
S.~Phithakkitnukoon, Z.~Smoreda, and P.~Olivier, ``Socio-geography of human
  mobility: A study using longitudinal mobile phone data,'' \emph{PloS one},
  vol.~7, no.~6, p. e39253, 2012.

\end{thebibliography}
\bibliographystyle{IEEEtran}

\end{document}